\documentclass[10pt]{article} %
\PassOptionsToPackage{table}{xcolor} %
\usepackage[preprint]{tmlr}

\usepackage{amsmath,amsfonts,bm}

\def\eqref#1{equation~\ref{#1}}

\def\1{\bm{1}}

\DeclareMathAlphabet{\mathsfit}{\encodingdefault}{\sfdefault}{m}{sl}
\SetMathAlphabet{\mathsfit}{bold}{\encodingdefault}{\sfdefault}{bx}{n}

\usepackage{url}

\newcommand{\dialogue}[1]{\noindent\textbf{#1}: }

\newcommand{\ours}[1]{BIGFix}

\newcommand{\TODO}[1]{\textbf{\color{red}[TODO: #1]}}

\renewcommand{\TODO}[1]{}

\newcommand{\rev}[1]{#1}

\definecolor{iccvblue}{rgb}{0.21,0.49,0.74}
\definecolor{darkorange}{RGB}{200,100,0}
\definecolor{darkviolet}{RGB}{148,0,211}
\newcommand{\imagenetid}[1]{\texttt{\textbf{#1}}}
\usepackage[pagebackref,breaklinks,colorlinks,allcolors=iccvblue]{hyperref}
\usepackage{amssymb}
\usepackage{amsfonts}
\usepackage{pifont} %
\usepackage{graphicx}
\usepackage{booktabs}
\usepackage{multirow}
\usepackage{adjustbox}
\usepackage{natbib}
\usepackage{subcaption}
\usepackage{soul}
\usepackage{booktabs} %
\usepackage{enumitem}
\usepackage{algorithm}
\usepackage{algpseudocode}
\usepackage{tikz}
\usetikzlibrary{arrows.meta, positioning, decorations.pathreplacing}

\title{BIGFix: Bidirectional Image Generation with Token Fixing}

\author{\centering
  Victor Besnier$^{2}$ \quad
  David Hurych$^{1}$ \quad
  Andrei Bursuc$^{2}$ \quad
  Eduardo Valle$^{2}$ 
  \\
  $^{1}$Valeo.ai, Prague, Czech Republic \quad
  $^{2}$Valeo.ai, Paris, France \\
  \textit{Corresponding author:} \texttt{\{victor.besnier\}@valeo.com} 
}

\def\month{MM}  %
\def\year{YYYY} %
\def\openreview{\url{https://openreview.net/forum?id=XXXX}} %

\begin{document}

\maketitle

\begin{figure}[h]
  \centering
  \includegraphics[width=\linewidth]{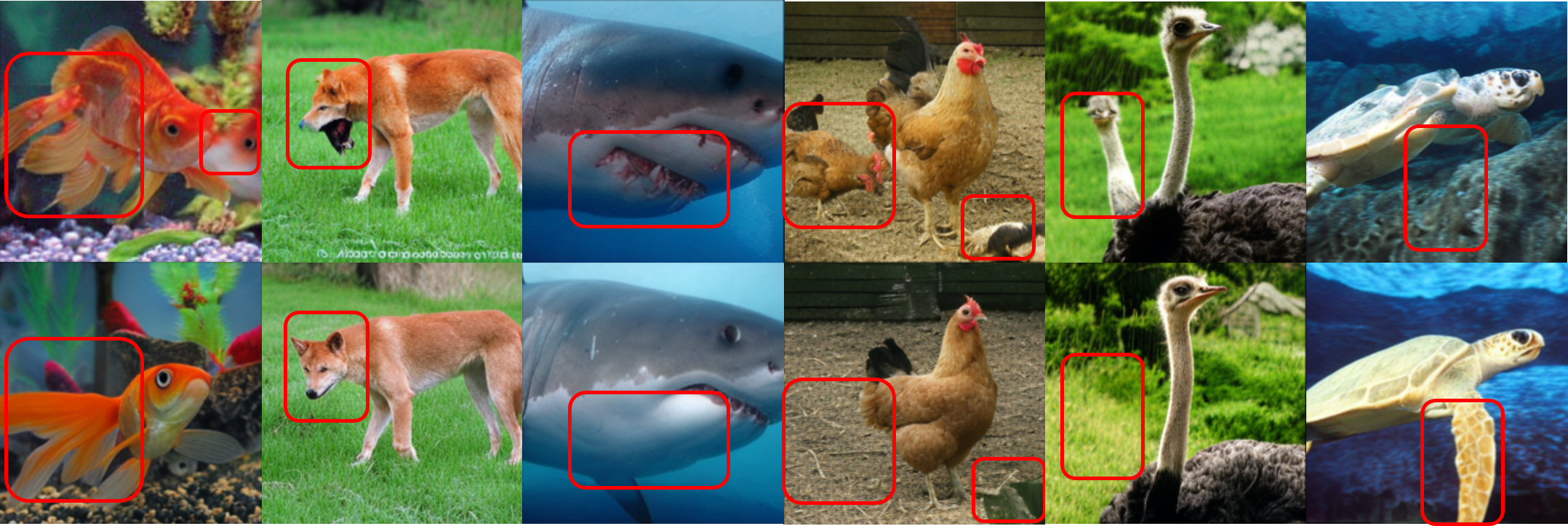}
   \caption{\textbf{Self-correction on ImageNet:} After the first 6 unmasking steps to create the overall structure, we proceed with two generations: without correction (top) and ours with correction (bottom). Correction solves structural errors such as supernumerary or missing elements.} 
   \label{fig:self_correction}
\end{figure}

\begin{abstract}
    
    Recent advances in image and video generation have raised significant interest from both academia and industry.
    A key challenge in this field is to improve both inference efficiency and quality, as model size and the number of inference steps directly impact the commercial viability of generative models while also posing fundamental scientific challenges.
    A promising direction involves combining auto-regressive sequential token modeling with multi-token prediction per step, reducing inference time by up to an order of magnitude.
    However, predicting multiple tokens in parallel can introduce structural inconsistencies due to token incompatibilities, as capturing complex joint dependencies during training remains challenging.
    \rev{Standard multi-token samplers keep every token once it is sampled, and existing correction methods rely on an auxiliary critic network or a separate correction stage.
    We propose \ours{}, which revises previously sampled tokens in the same forward pass that predicts the next ones, without any auxiliary network.
    During training, a fraction of the context tokens is replaced by tokens swapped from the same image, a corruption that preserves the token distribution of the image, and the model learns to both restore the context and predict the next tokens.}
    Our method preserves the efficiency benefits of parallel token prediction while significantly enhancing generation quality.
    We evaluate our approach on class-to-image (ImageNet-256, CIFAR-10), text-to-image (GPIC), class-to-video (UCF-101) and image-to-video (NuScenes), demonstrating substantial improvements across multiple tasks. Code and pre-trained models are available at \url{https://github.com/valeoai/BigFix}.
    
\end{abstract}

\section{Introduction}
Surrounded by the aroma of freshly brewed coffee, casual chat in the office shifts from everyday topics to the latest breakthroughs in image generation. As usual, the discussion quickly shifts towards the constant push for state-of-the-art models in computer vision, reflecting the rapid evolution in the field and the drive to refine its techniques.

\dialogue{Eritos\footnote{fictive name (from erōtáō, “to ask”)}} “Have you seen the news from BlackForest~\citep{flux2024}? The quality is mesmerizing, indistinguishable from reality! Far superior to those chicken images you've been generating for months."

\dialogue{Theoros\footnote{fictive name (from theōrein, “to seek")}} “Indeed, I saw it! They used flow matching, scaling it across many images, parameters, and GPUs. It's impressive how they managed to push the boundaries. But, for the record, my generated chickens are also quite remarkable! (\autoref{fig:chicken})"

\dialogue{E} “Flow matching? Again a new framework? Is it related to diffusion model?"

\dialogue{T} “Flow matching~\citep{lipman2023flow} is a method that teaches the model to transform a simple distribution, like a Gaussian, into a complex one, such as an image, by following smooth, continuous paths. Flow matching and diffusion~\citep{ho2020ddpm} currently lead image generation, surpassing traditional methods like GAN~\citep{goodfellow2020gan} or VAE~\citep{kingma2013vae}."

\dialogue{E} “I thought that auto-regressive model, like GPT~\citep{radford2019language}, was SOTA for data generation. I know text is discrete but an image is also a collection of discrete pixel values from 0 to 255, right? Why can’t you just generate a pixel like you generate a word?"

\dialogue{T} “Technically, you can~\citep{chen2020igpt}, but only for very small images. And for a $256\times256$ image, you’d need 65,536 iterations to generate just one sample... and that’s assuming a single RGB value per pixel. Then you have three color channels, so the vocabulary size is $256^{3}$. Imagine the number of parameters and the compute power needed for training and inference. It’s a nightmare."

\dialogue{E} “Indeed. But scaling usually works~\citep{hoffmann2022training}, right?"

\dialogue{T} “Sure, but that's not all. Images are inherently 2D structures, yet auto-regressive methods ~\citep{sun2024autoregressive} often enforce a raster-scan ordering that ignores spatial organization. The notion of geometric neighborhood is very meaningful for pixels. And this is different in text, where semantics and positional distance are much less correlated~\citep{tian2024visual}.

\dialogue{E} “Okay, but what about compressing both the image resolution and its range of values?"

\dialogue{T} “Well, yes. Techniques using vector quantization~\citep{esser2020taming} tokenize images, reducing their resolution and converting them into a discrete set of tokens~\citep{menter2024fsq,yu2024magvit2}. But synthesizing images token-by-token remains computationally expensive even when predicting in the latent space. Moreover, causal attention, which works well for text, is not ideal for images, as a basic raster-scan order fails to capture the conditional structure of images."

\dialogue{E} “So, what's the alternative? Can't we generate tokens in a different order?"

\dialogue{T} “This brings us to Masked Generative Image Transformer~\citep{chang2022maskgit,chang2023muse}! These models are trained with a reconstruction objective to predict all tokens at once. During inference, the model progressively fills a fully-masked image by predicting multiple tokens in parallel without fixing the order. This framework accelerates the image generation process, producing results in just a few steps."

\dialogue{E} “Amazing! Then why isn't everyone using this?"

\dialogue{T} “There are a couple of challenges. MaskGIT iteratively reveals image tokens, which can lead to sampling discrepancies when incompatible tokens are independently sampled simultaneously~\citep{lezama2022tokencritic}. \rev{Moreover, the standard MaskGIT sampler never revisits a token: once sampled, it stays forever. Some methods train a separate critic to remask doubtful tokens, but that is a second network to train and run at every step. Unless the generator itself learns to fix its mistakes."}

\dialogue{\rev{E}} “Interesting. Sounds like a problem worth solving, right?!"

\paragraph{Contributions.} \rev{Training on corrupted inputs and revising generated tokens both have precedents: BERT randomly replaces input tokens~\citep{devlin2019bert}, discrete diffusion uses uniform corruption kernels~\citep{austin2021structured}, SUNDAE trains on its own corrupted samples~\citep{savinov2022stepunrolled}, and Token-Critic, discrete predictor-corrector models and ReMDM revise already sampled tokens~\citep{lezama2022tokencritic, lezama2023discrete, wang2025remasking} (see~\autoref{sec:RelatedWork}). This paper brings self-correction to multi-token visual generation at no extra inference cost. Our contributions are:
\textbf{(1)} An in-image corruption for training: a fraction $\alpha$ of the context tokens is replaced by tokens swapped from the same image, which preserves the token distribution of the image while breaking its spatial consistency. The network learns both to restore the corrupted context and to predict the next group of tokens, with $\alpha$ as the only additional hyper-parameter.
\textbf{(2)} Correction in the same forward pass as next-group prediction: during sampling, no tokens are injected; instead, the model re-predicts every previously sampled token while it predicts the next group (\autoref{fig:scheme}). This requires no auxiliary critic, no extra correction stage and no additional forward pass.
\textbf{(3)} A controlled evaluation across class-to-image, text-to-image and video generation, including an ablation of the corruption kernel, a matched comparison with inference-time remasking, and a compositional benchmark.}

\begin{figure}
    \centering
    \resizebox{\linewidth}{!}{\definecolor{tkMask}{RGB}{226,229,234}
\definecolor{tkCtx}{RGB}{106,150,205}
\definecolor{tkNext}{RGB}{245,166,80}
\definecolor{tkErr}{RGB}{217,83,79}
\definecolor{tkFix}{RGB}{92,174,100}
\definecolor{tkCard}{RGB}{246,247,249}
\begin{tikzpicture}[
    font=\footnotesize,
    cmask/.style={fill=tkMask, rounded corners=0.6pt},
    cctx/.style={fill=tkCtx, rounded corners=0.6pt},
    cnxt/.style={fill=tkNext, rounded corners=0.6pt},
    ctgt/.style={fill=tkMask, draw=tkNext, line width=0.7pt, rounded corners=0.6pt},
    cerr/.style={fill=tkErr, rounded corners=0.6pt},
    cfix/.style={fill=tkFix, rounded corners=0.6pt},
    flow/.style={-{Stealth[length=4pt, width=3.5pt]}, line width=0.7pt, black!55},
    swap/.style={-{Stealth[length=3pt]}, densely dashed, line width=0.5pt, tkErr},
    title/.style={anchor=west, font=\small\bfseries, text=black!85},
    sub/.style={font=\scriptsize, text=black!65, align=center},
]
\newcommand{\tkgrid}[5]{%
    \begin{scope}[shift={(#1,#2)}]
        \foreach \i in {0,...,5}{\foreach \j in {0,...,5}{%
            \path[#4] (\i*#3,-\j*#3) rectangle ++(0.84*#3,-0.84*#3);}}
        \foreach \i/\j/\s in {#5}{\path[\s] (\i*#3,-\j*#3) rectangle ++(0.84*#3,-0.84*#3);}
    \end{scope}}

\fill[tkCard, rounded corners=5pt] (0,0.45) rectangle (16.0,-3.05);
\node[title] at (0.15,0.15) {(a) Training: in-image token swap};

\tkgrid{0.5}{-0.25}{0.34}{cmask}{1/1/cctx,4/1/cctx,2/4/cctx,5/4/cctx,0/3/cctx,3/0/cctx,5/2/cctx,2/2/cctx,3/5/cctx,0/5/cctx,1/0/ctgt,5/0/ctgt,3/3/ctgt,1/5/ctgt,4/3/ctgt}
\node[sub] at (1.5,-2.68) {clean context $\mathcal{Z}_{<s}$\\and next group $\mathcal{Z}_s$};

\draw[flow] (2.75,-1.27) -- (3.75,-1.27) node[midway, above, font=\footnotesize]{$\Phi_\alpha$};

\tkgrid{4.0}{-0.25}{0.34}{cmask}{1/1/cctx,4/1/cctx,2/4/cctx,5/4/cerr,0/3/cctx,3/0/cctx,5/2/cctx,2/2/cerr,3/5/cctx,0/5/cctx,1/0/ctgt,5/0/ctgt,3/3/ctgt,1/5/ctgt,4/3/ctgt}
\draw[swap] (4.143,-1.413) to[bend left=40] (4.823,-1.073);
\draw[swap] (5.163,-2.093) to[bend right=40] (5.843,-1.753);
\node[sub] at (5.0,-2.68) {corrupted input $\widetilde{\mathcal{Z}}_{<s}$\\(swaps within the image)};

\draw[flow] (6.25,-1.27) -- (7.05,-1.27);
\node[draw=black!25, fill=white, rounded corners=4pt, minimum width=1.75cm, minimum height=1.35cm, align=center]
    at (7.95,-1.27) {$P_\theta$\\[1pt]{\scriptsize\color{black!60}bidirectional}\\[-2pt]{\scriptsize\color{black!60}transformer}};
\draw[flow] (8.85,-1.27) -- (9.65,-1.27);

\tkgrid{9.9}{-0.25}{0.34}{cmask}{1/1/cctx,4/1/cctx,2/4/cctx,5/4/cfix,0/3/cctx,3/0/cctx,5/2/cctx,2/2/cfix,3/5/cctx,0/5/cctx,1/0/cnxt,5/0/cnxt,3/3/cnxt,1/5/cnxt,4/3/cnxt}
\node[sub] at (10.9,-2.68) {prediction $\widehat{\mathcal{Z}}_{\le s}$\\(one forward pass)};

\begin{scope}[shift={(12.35,-0.75)}]
    \path[cfix] (0,0) rectangle ++(0.24,-0.24);
    \node[anchor=west, align=left] at (0.3,-0.12) {$\mathcal{L}_{\mathrm{context}}$: restore\\the swapped tokens};
    \path[cnxt] (0,-0.9) rectangle ++(0.24,-0.24);
    \node[anchor=west, align=left] at (0.3,-1.02) {$\mathcal{L}_{\mathrm{next}}$: predict\\the next group $\mathcal{Z}_s$};
\end{scope}

\fill[tkCard, rounded corners=5pt] (0,-3.25) rectangle (16.0,-8.05);
\node[title] at (0.15,-3.55) {(b) Sampling: correction in the same forward pass};

\foreach \x/\l in {3.4/step $1$, 6.3/step $2$, 9.2/step $3$, 12.1/last step $m$}{\node[sub] at (\x+0.9,-3.85) {\l};}
\foreach \y in {-4.05,-6.0}{%
    \foreach \x in {5.35,8.25}{\draw[flow] (\x,\y-0.9) -- (\x+0.8,\y-0.9);}
    \draw[flow, densely dotted, line width=0.9pt] (11.15,\y-0.9) -- (11.95,\y-0.9);
    \node[font=\scriptsize, text=black!55] at (11.55,\y-0.68) {$\cdots$};}

\node[anchor=west, align=left] at (0.3,-4.95) {\textbf{Halton-MaskGIT}\\[1pt]{\scriptsize\color{black!60}no correction}};
\tkgrid{3.4}{-4.05}{0.3}{cmask}{1/1/cnxt,4/1/cerr,2/4/cnxt,5/4/cnxt}
\tkgrid{6.3}{-4.05}{0.3}{cmask}{1/1/cctx,4/1/cerr,2/4/cctx,5/4/cctx,0/3/cnxt,3/0/cnxt,5/2/cerr,2/2/cnxt,3/5/cnxt,0/5/cerr}
\tkgrid{9.2}{-4.05}{0.3}{cmask}{1/1/cctx,4/1/cerr,2/4/cctx,5/4/cctx,0/3/cctx,3/0/cctx,5/2/cerr,2/2/cctx,3/5/cctx,0/5/cerr,1/0/cnxt,5/0/cnxt,3/3/cnxt,1/5/cnxt,4/3/cerr,0/1/cnxt,2/0/cnxt,5/5/cnxt}
\tkgrid{12.1}{-4.05}{0.3}{cctx}{0/0/cnxt,4/0/cnxt,2/1/cnxt,3/1/cnxt,5/1/cnxt,0/2/cnxt,1/2/cnxt,3/2/cnxt,4/2/cnxt,1/3/cnxt,2/3/cnxt,5/3/cnxt,0/4/cnxt,1/4/cnxt,3/4/cnxt,4/4/cnxt,2/5/cnxt,4/5/cnxt,4/1/cerr,5/2/cerr,0/5/cerr,4/3/cerr}
\node[anchor=west, sub, text=tkErr, align=left] at (14.05,-4.95) {errors kept\\until the end};

\node[anchor=west, align=left] at (0.3,-6.9) {\textbf{\ours{}}\\[1pt]{\scriptsize\color{black!60}with correction}};
\tkgrid{3.4}{-6.0}{0.3}{cmask}{1/1/cnxt,4/1/cerr,2/4/cnxt,5/4/cnxt}
\tkgrid{6.3}{-6.0}{0.3}{cmask}{1/1/cctx,4/1/cfix,2/4/cctx,5/4/cctx,0/3/cnxt,3/0/cnxt,5/2/cnxt,2/2/cnxt,3/5/cnxt,0/5/cnxt}
\tkgrid{9.2}{-6.0}{0.3}{cmask}{1/1/cctx,4/1/cfix,2/4/cctx,5/4/cctx,0/3/cctx,3/0/cctx,5/2/cctx,2/2/cctx,3/5/cctx,0/5/cctx,1/0/cnxt,5/0/cnxt,3/3/cnxt,1/5/cnxt,4/3/cnxt,0/1/cnxt,2/0/cnxt,5/5/cnxt}
\tkgrid{12.1}{-6.0}{0.3}{cctx}{0/0/cnxt,4/0/cnxt,2/1/cnxt,3/1/cnxt,5/1/cnxt,0/2/cnxt,1/2/cnxt,3/2/cnxt,4/2/cnxt,1/3/cnxt,2/3/cnxt,5/3/cnxt,0/4/cnxt,1/4/cnxt,3/4/cnxt,4/4/cnxt,2/5/cnxt,4/5/cnxt,4/1/cfix,3/3/cfix,1/5/cfix}
\node[anchor=west, sub, text=tkFix!80!black, align=left] at (14.05,-6.9) {errors fixed\\at every step};

\begin{scope}[shift={(0.5,-8.35)}]
    \foreach \x/\s/\l in {0/cmask/masked token, 2.5/cctx/context token, 5.0/cnxt/next group (newly sampled), 9.0/cerr/erroneous token, 11.8/cfix/corrected token}{%
        \path[\s] (\x,0.12) rectangle ++(0.24,-0.24);
        \node[anchor=west, font=\scriptsize, text=black!75] at (\x+0.3,0) {\l};}
\end{scope}
\end{tikzpicture}}
    
    \caption{\textbf{Schematics.} \rev{(a) Training: a fraction $\alpha$ of the context $\mathcal{Z}_{<s}$ is swapped with tokens of the same image (red); in one forward pass, the model restores them ($\mathcal{L}_{\mathrm{context}}$, green) and predicts the next group $\mathcal{Z}_s$ ($\mathcal{L}_{\mathrm{next}}$, orange). (b) Sampling: Halton-MaskGIT keeps every sampled token, so an incompatible token (red) stays until the end; \ours{} re-predicts the whole context at every step, the last one included, and corrects it (green).}
    }
    \label{fig:scheme}
\end{figure}

\section{Related work}\label{sec:RelatedWork}

\paragraph{Continuous Visual Generative Modeling.}
Generative image modeling has long intrigued researchers, and Generative Adversarial Networks (GANs)~\citep{goodfellow2020gan, kang2023scaling} have pioneered this effort, but suffer from mode collapse issues, which prevent them from covering the full distribution of the data~\citep{liu2023combating}. Diffusion Models have emerged as leading approach in generative modeling~\citep{song2021denoising, dhariwal2021diffusion}. They are trained in two phases: noise is progressively added to images, and a model is learned to reverse this process via denoising. Since their introduction, diffusion models have achieved strong results in both text-to-image~\citep{ramesh2022hierarchicaltextconditionalimagegeneration, rombach2022ldm} and text-to-video~\citep{ho2022videodiffusionmodels, singer2022makeavideotexttovideogenerationtextvideo} generation. Building on Diffusion, Flow matching~\citep{lipman2023flow} offers an alternative formulation. Using continuous normalizing flows, it efficiently transforms noise into data distributions. It has since been successfully applied to both videos~\citep{jin2024pyramidal} and images~\citep{esser2024scaling}.

\paragraph{Auto-Regressive Visual Modeling.}
Inspired by advances in natural language processing~\citep{radford2019language, devlin2019bert}, auto-regressive and masked generative methods have been adapted for image and video synthesis. iGPT~\citep{chen2020igpt} modeled images as pixel sequences auto-regressively, though quadratic scaling of Transformers limited its use to small images only. To overcome this issue, VQ-GAN~\citep{esser2020taming} employs a vector-quantized variational auto-encoder\citep{van2017vqvae} with perceptual and adversarial losses, compressing images into a grid of discrete tokens for realistic reconstruction. ViT-VQGAN\citep{yu2022vectorquantized} increases scalability with Vision Transformers, while VideoGPT~\citep{yan2021videogpt} extends the approach to video generation. DALL-E, Parti and LlamaGen~\citep{ramesh2021dalle, yu2022parti, sun2024autoregressive} proposed further adaptations for the text-to-image synthesis. 

\paragraph{Masked Visual Modeling.}
Inspired by BERT~\citep{devlin2019bert}, MaskGIT~\citep{chang2022maskgit, chang2023muse} introduced an alternative to auto-regressive modeling. Challenging the slow, row-wise token generation of auto-regressive methods, MaskGIT adopts bidirectional prediction. Trained to uncover randomly masked image tokens, it begins inference with a fully masked image, iteratively unmasking tokens to produce a full image. At each step, MaskGIT predicts all tokens but retains only the most confident ones while the remaining stay masked. MaskGIT demonstrates decent quality of generated images and up to an order of magnitude faster inference over auto-regressive models~\citep{villegas2022phenaki, yu2023magvitmaskedgenerativevideo, yu2024magvit2}. However, parallel token sampling overlooks inter-token dependencies, potentially yielding suboptimal results, prompting alternative sampling techniques~\citep{lezama2022tokencritic, lee2023enhancedsampling, besnier2025halton} to address this limitation.

\rev{\paragraph{Corrupted Context and Token Correction.}
Training on corrupted inputs is a long-standing idea. BERT~\citep{devlin2019bert} replaces a small fraction of the masked tokens with random vocabulary tokens to learn representations, not to generate. Discrete diffusion models such as D3PM~\citep{austin2021structured} define forward processes with uniform transition kernels, in which any token can be replaced by any other codebook entry, and SUNDAE~\citep{savinov2022stepunrolled} trains a text denoiser on its own corrupted samples. Several methods revise tokens that are already generated. Token-Critic~\citep{lezama2022tokencritic} and the enhanced sampling scheme of~\citet{lee2023enhancedsampling} train an auxiliary critic that selects the tokens to remask, discrete predictor-corrector diffusion~\citep{lezama2023discrete} learns a separate corrector, and ReMDM~\citep{wang2025remasking} re-masks generated tokens at inference time without retraining. \ours{} differs in three ways: correction happens inside the forward pass that already predicts the next tokens, it requires neither an auxiliary network nor an extra stage, and the corruption swaps tokens within the same image instead of drawing them from the whole codebook, which keeps the injected tokens plausible.}

\paragraph{Multi-token Auto-regressive Methods.}
To address the limitations of auto-regressive and masked generative modeling, Visual AutoRegressive Modeling (VAR)~\citep{tian2024visual} introduces a coarse-to-fine next-scale prediction strategy, auto-regressively modeling multi-scale token maps from low to high detail, while still leveraging parallel token prediction at each scale. Another line of work includes methods such as ZIPAR~\cite{he2025zipar} and NAR~\citep{He_2025_ICCV}, which accelerate generation by decoding local windows of neighbor tokens in parallel, achieving substantial speedups without significantly compromising image quality. \rev{ARPG~\citep{li2026autoregressive} decodes tokens in a randomized order with parallel predictions, and DetailFlow~\citep{liu2025detailflow} generates a coarse-to-fine 1D token sequence and adds a self-correction mechanism for tokens sampled in parallel. Both modify the tokenization or the decoding order, whereas \ours{} keeps the masked-transformer formulation and only changes the training inputs and the sampling update.}

\paragraph{Auxiliary Training Strategies.} In this study, we exclude auxiliary losses such as REPA~\citep{yu2026representation}, or reinforcement learning finetuning~\citep{wallace2024diffusion} which improve the quality of the generator. Similar for model distillation training~\citep{salimans2022progressive} which reduces the number of steps. While effective, and applicable in tandem with our method, they typically require either additional pre-trained networks during inference or a separate fine-tuning stage on top of the vanilla approach. However, to ensure a fair comparison against modern baselines, we apply reward-conditioned~\citep{dufour2024dont,dufour2025miro} post-training exclusively to our final Text-to-Image models.

Here, we want to keep the advantage of fast inference time and to combine multi-token and auto-regressive approaches. Therefore our baseline method is \cite{besnier2025halton}. \rev{The goal is to improve multi-token discrete generation, not to compete with the largest diffusion and flow models.} In contrast to the above mentioned methods, we equip the model with the ability to recover from erroneous token predictions at each prediction step. Our random token injection at training time and ability to fix errors at inference time is well suited for any multi-token prediction technique and may be easily incorporated into these frameworks.

\section{Method}\label{sec:Method}
\subsection{Preliminaries: Multi-Token Prediction}
\paragraph{Tokenizing Images.}
We represent each image \( x  \in \mathbb{R}^{H \times W \times 3} \) as a set of discrete tokens \( \mathcal{Z}=\{z_{0}, z_{1}, \dots, z_{n}\} \) using a pre-trained tokenizer encoder \( \text{Enc} \):  
\[
\mathcal{Z} = \text{Enc}(x), \quad \mathcal{Z} \in C^{h \times w}
\]
where \( C \) denotes the token vocabulary, i.e., the possible values each \(z_{i}\) can take, and \( h \times w \) represents the spatial dimensions of the token grid. 
The main objective in image synthesis is to learn how to sample from the data distribution \(P(\mathcal{Z})\). For convenience, we can factorize \(P(\mathcal{Z})\) as the joint distribution over all tokens:  
\begin{equation}
P(\mathcal{Z}) = P(z_{0}, \dots, z_{n}).
\end{equation}

\paragraph{Auto-regressive Modeling.}  
A powerful tool to learn the data distribution, heavily used to train LLMs, is to represent the data as a sequence and learn the probability of the next token given the previous ones (the context). Auto-regressive approaches estimate the conditional probability of each token based on the previously generated tokens:
\begin{equation}
    P_{\theta}(z_{i} \mid z_{0}, z_{1}, \dots, z_{i-1}),
\end{equation}

where \( \theta \) represents the model parameters. 
During generation, auto-regressive methods sequentially sample each token by leveraging the product of these probabilities:
\begin{equation}
P(z_{0}, \dots, z_{n}) = \prod\limits_{i=0}^{n} P(z_{i} \mid z_{0}, z_{1}, \dots, z_{i-1}).
\end{equation}

Here, \( n = h \times w \) is the total number of tokens. By construction, auto-regressive models enforce sequential token generation, preventing the simultaneous sampling of multiple tokens. This constraint results in a computational bottleneck, requiring \( n \) forward passes, which is prohibitively expensive for images (e.g., \( 1024 \) steps for a small \( 32 \times 32 \) image). Furthermore, this approach forces to flatten the 2D spatial structure of image tokens into a raster-scan ordering in 1D. This transformation corrupts the inherent spatial relationships of tokens (See~\autoref{sec:sampling_patern}). 

\paragraph{Multi-token prediction.}   %
To mitigate the inefficiency of auto-regressive methods, we redefine the image representation as an ordered sequence of groups of tokens \( \mathcal{S} = \{\mathcal{Z}_0, \mathcal{Z}_1, \dots, \mathcal{Z}_m \} \), where each subset \( \mathcal{Z}_i \) is a non-empty set of tokens selected from \( \{z_i\}_{i=0}^{n} \). All subsets \( \mathcal{Z}_s \) form a non-overlapping and complete partitioning of \( \{z_i\}_{i=0}^{n} \) . Instead of predicting one token at a time, we model the conditional probability of entire token groups \( P_{\theta}(\mathcal{Z}_{s} \mid \mathcal{Z}_{0}, \mathcal{Z}_{1}, \dots, \mathcal{Z}_{s-1}) \) resulting in the factorized probability distribution $P(\mathcal{S})$:
\begin{equation}
P(\mathcal{Z}_0, \mathcal{Z}_1, \dots, \mathcal{Z}_m) = \prod\limits_{s=0}^{m} P(\mathcal{Z}_{s} \mid \mathcal{Z}_{0}, \mathcal{Z}_{1}, \dots, \mathcal{Z}_{s-1}).
\end{equation}

Setting \( m \ll n \), i.e., reducing the number of sampling steps, significantly accelerates the generation process. Each step \( s \) predicts a set of next tokens \( \mathcal{Z}_s \). Moreover, the model can still be trained using the standard cross-entropy loss.

\subsection{Toward Self-Correcting Token Generation}
\paragraph{On token dependencies.}
However, a major drawback of sampling multiple tokens simultaneously is its sensitivity to errors. \rev{The network outputs are not independent: all positions share the same context through attention. The independence comes from the sampling step, which draws every token of a group separately given the context $\mathcal{C} = \{\mathcal{Z}_0, \dots, \mathcal{Z}_{s-1}\}$:
\[
P(z_a, z_b \mid \mathcal{C}) \approx P_{\theta}(z_a \mid \mathcal{C})\, P_{\theta}(z_b \mid \mathcal{C}),
\]

where \( P(z_a, z_b \mid \mathcal{C}) \) is the true joint probability of two tokens \( z_a, z_b \) of the same group, and the right-hand side is the product of the per-position predictions, trained with a cross-entropy loss on each position.} In reality, token dependencies in images often cause situations where individual tokens may have high probabilities, yet their joint probability is significantly lower, leading to unrealistic or inconsistent generations~\citep{besnier2025halton} 
\footnote{For example, in a picture, a person's head might appear in two different locations. While each position may have a high likelihood for the token corresponding to the  ‘head’ independently, selecting both simultaneously could result in a person with two heads—an improbable and unrealistic outcome. Such errors may propagate through the sampling process, leading to inconsistencies in predictions.}.

\rev{This conditional independence at sampling time is} sufficient to create compositional problems, which we demonstrate quantitatively in the next section and qualitatively~\autoref{fig:self_correction}.

\paragraph{Random token injection.}

To mitigate this issue and enable the model to recover from incompatible context tokens, we corrupt the input context during training by randomly replacing a subset of its tokens. Given a tokenized image
$\mathcal Z=\{z_i\}_{i=1}^{N}$,
we define a corruption operator
$\Phi_{\alpha}$,
parameterized by the injection probability $\alpha$, which produces a corrupted token sequence
$\widetilde{\mathcal Z}=\Phi_{\alpha}(\mathcal Z)$.
For every context token, we independently sample

\begin{equation}
\tilde z_i =
\begin{cases}
z_i, & u_i>\alpha,\\
z_j,\quad j\sim\mathcal U(\{1,\ldots,N\}), & \text{otherwise},
\end{cases}
\qquad
u_i\sim\mathcal U(0,1).
\label{eq:random_injection}
\end{equation}

That is, with probability $\alpha$, a context token is replaced by another token sampled uniformly from the same clean image. Since the injected tokens originate from the same image, they remain semantically plausible while introducing inconsistencies in the local context. Training on $\widetilde{\mathcal Z}$ therefore encourages the model to identify and correct corrupted context tokens instead of relying on them blindly. \rev{Unlike the uniform kernels of BERT or discrete diffusion~\citep{devlin2019bert, austin2021structured}, which draw replacements from the whole codebook, this in-image swap preserves the token distribution of the image: the injected tokens are plausible, only their position is wrong, which mimics the structural errors of parallel sampling. \autoref{sec:kernel} compares this kernel with uniform codebook replacement and with tokens sampled by the model itself.}

\paragraph{Model Training.} We inject random token and train our model to correct them by maximizing the likelihood of \(P_{\theta}(\mathcal{Z}_{0}, \dots, \mathcal{Z}_{k} \mid\widetilde{\mathcal Z}_{0}, \dots, \widetilde{\mathcal Z}_{k})\) which corresponds to minimizing : 
\begin{equation}
    \mathcal{L}_{context}(\theta) = - \mathbb{E}_{\underset{\mathcal{Z}, \tilde{\mathcal Z}}{}} \left[\sum_{k=0}^{s-1} \log P_{\theta}(\mathcal{Z}_{0}, \dots, \mathcal{Z}_{k} \mid \widetilde{\mathcal Z}_{0}, \dots, \widetilde{\mathcal Z}_{k})\right].
    \label{eq:l_context}
\end{equation}
 
where $k < s$ ensures that we inject random tokens only in the context.

Moreover, we train the model to predict the next clean token group \(\mathcal{Z}_{s}\) given a sequence of noisy token groups \( \{\widetilde{\mathcal Z}_{0}, \dots, \widetilde{\mathcal Z}_{s-1}\}\), thereby making it a generative model. We minimize the expected negative log-likelihood with respect to \(\theta\):
\begin{equation}
\mathcal{L}_{next}(\theta) = - \mathbb{E}_{\underset{\mathcal{Z}, \tilde{\mathcal Z}}{}} \left[\sum_{s=0}^{m}\log P_{\theta}(\mathcal{Z}_s \mid \widetilde{\mathcal Z}_{0}, \widetilde{\mathcal Z}_{1}, \dots, \widetilde{\mathcal Z}_{s-1}) \right].
\end{equation}

In practice, we train the model to minimize the sum $\mathcal{L}(\theta)~=~\mathcal{L}_{next}(\theta)~+~\mathcal{L}_{context}(\theta)$ (\autoref{alg:train_self_correcting}). During sampling, no injections are made into the context; instead, the model iteratively predict the next tokens and correct its previous steps in a single forward pass (\autoref{alg:sample_self_correcting}).

\rev{\paragraph{Applying the corrections.} At every step, every context token is overwritten by a sample from the model's prediction at its position, with the same temperature, top-$p$ and classifier-free guidance as the newly predicted tokens. There is no confidence threshold and no partial update. Since a bidirectional transformer already outputs a distribution for every position, the correction costs no extra forward pass; the logits that a standard masked sampler discards are used instead. In practice, no oscillation appears: the model reproduces the tokens it is confident about, and about 10\% of the final tokens change over the whole trajectory (\autoref{sec:token_fixing}).}

\paragraph{Bidirectional Halton ordering.}%
Building on prior work~\citep{besnier2025halton}, which serves as our main baseline method, we construct the set \( \mathcal{S} = \{\mathcal{Z}_0, \mathcal{Z}_1, \dots, \mathcal{Z}_m \} \) using the Halton sequence to determine the prediction order of each token \( z_{i} \). The Halton sequence, a low-discrepancy sequence, ensures uniform token coverage across the 2D spatial structure of the image while reducing the mutual information shared within each set. Consequently, we retain the bidirectional attention mechanism, which leverages the 2D nature of images to facilitate global contextual understanding, akin to Transformer encoder-style models. Finally, we adopt an arccos scheduling scheme to progressively increase the number of tokens within each group \( \mathcal{Z}_s \), thereby balancing uniform token distribution with efficient sampling dynamics.

\begin{figure}[t]
\small
\begin{minipage}[t]{0.495\linewidth}
\begin{algorithm}[H]
\caption{Self-Correcting Multi-Token Training}
\label{alg:train_self_correcting}
\begin{algorithmic}[1]
\Require groups $\mathcal{S}=\{\mathcal{Z}_0,\ldots,\mathcal{Z}_m\}$, model $P_\theta$, corruption $\Phi_{\alpha}$
\For{each minibatch}
    \State $s \sim \mathcal{U}(\{0,\ldots,m\})$ \Comment{random step}
    \State $\widetilde{\mathcal Z}_{<s} \gets \Phi_{\alpha}(\mathcal Z_{<s})$ \Comment{corrupt context}
    \State $\widehat{\mathcal Z}_{\le s} \gets P_\theta(\widetilde{\mathcal Z}_{<s})$ \Comment{one pass}
    \State $\mathcal L_{\mathrm{next}} \gets \mathrm{CE}(\widehat{\mathcal Z}_{s},\mathcal Z_{s})$ \Comment{next group}
    \State $\mathcal L_{\mathrm{context}} \gets \mathrm{CE}(\widehat{\mathcal Z}_{<s},\mathcal Z_{<s})$ \Comment{context}
    \State Update $\theta$ with $\nabla_\theta (\mathcal L_{\mathrm{next}} + \mathcal L_{\mathrm{context}})$
\EndFor
\end{algorithmic}
\end{algorithm}
\end{minipage}\hfill
\begin{minipage}[t]{0.495\linewidth}
\begin{algorithm}[H]
\caption{Self-Correcting Multi-Token Sampling}
\label{alg:sample_self_correcting}
\begin{algorithmic}[1]
\Require model $P_\theta$, groups $\mathcal{S}=\{\mathcal{Z}_0,\ldots,\mathcal{Z}_m\}$
\State $\widehat{\mathcal{Z}} \gets$ mask tokens
\For{$s = 0, \dots, m$}
    \State $\widehat{\mathcal{Z}}'_{<s},\, \widehat{\mathcal{Z}}'_s \gets P_\theta(\widehat{\mathcal{Z}}_{<s})$ \Comment{one pass}
    \State $\widehat{\mathcal{Z}}_{<s} \gets \widehat{\mathcal{Z}}'_{<s}$ \Comment{correct context}
    \State $\widehat{\mathcal{Z}}_{s} \gets \widehat{\mathcal{Z}}'_{s}$ \Comment{add new group}
\EndFor
\State \Return $\widehat{\mathcal{Z}}$
\end{algorithmic}
\end{algorithm}
\end{minipage}
\end{figure}

\section{Experiments}\label{sec:Results}

\subsection{Implementation details}

\paragraph{Model Architecture.}
Our models are based on the same repository as~\citep{besnier2025halton} with minimal changes for tokenizer, modality and loss adaptation. We use AdaLN for class conditioning similar to the DiT~\citep{Peebles2023scalable} and a simple cross-attention for the text condition. The complete hyper-parameter details are in~\autoref{sec:hparam}.

\paragraph{Evaluation Metrics.}
To assess image generation quality, we use Fr\'echet Inception Distance (FID)~\citep{heusel2017gans}, Inception Score (IS) and the Precision and Recall. For video generation, we rely on Fr\'echet Video Distance (FVD)~\citep{unterthiner2019fvd}. For text to image, we evaluate our model using four automatic metrics: ImageReward~\citep{xu2023imagereward}, HPSv2~\citep{wu2023human}, Aesthetics~\citep{schuhmann2021laion}, and PickScore~\citep{kirstain2023pick}.

\paragraph{Modalities.}
For class-to-image generation on ImageNet, we use a pre-trained LlamaGen tokenizer~\citep{sun2024autoregressive}, with a downscale spatial factor of 16 and a codebook of 16,384 codes. On Cifar10~\citep{cifar10}, we do not use any learnable tokenizer. Instead, images are quantized, mapping each RGB pixel to a single token using the formula: \(R + G \cdot q + B \cdot q^2 \) with a codebook size of \(q^3\). We set \(q=16\), yielding a codebook size of 4,096. For text to image results, we reuse the LlamaGen tokenizer~\citep{sun2024autoregressive} specifically trained on LAION~\citep{schuhmann2021laion}. For class-to-video generation on UCF-101~\citep{ucf101}, we use OmniTokenizer~\citep{wang2024omnitokenizer} to encode 17-frame videos with a codebook of 8,192 codes. For NuScenes~\citep{caesar2020nuscenes}, each frame is tokenized independently using LlamaGen. Finally, we use a single frame as conditioning (image-to-video) and generate the following 16 frames auto-regressively, with a maximum context of 8 frames. 

\begin{table}
\centering
\resizebox{\textwidth}{!}{%
\begin{tabular}{c|cc|cc|cc|c|c}
\toprule
 Dataset    & \multicolumn{2}{c|}{\textbf{ImageNet}} & \multicolumn{2}{c|}{\textbf{ImageNet}} & \multicolumn{2}{c|}{\textbf{CIFAR-10}}  & \multicolumn{1}{c|}{\textbf{UCF-101}}  & \multicolumn{1}{c}{\textbf{NuScenes}}\\\midrule
 Model      & \multicolumn{2}{c|}{Small}    & \multicolumn{2}{c|}{Base}     & \multicolumn{2}{c|}{Small}  & \multicolumn{1}{c|}{Base}  & \multicolumn{1}{c}{Base}\\\midrule
 Tokenizer  & \multicolumn{2}{c|}{LlamaGen} & \multicolumn{2}{c|}{LlamaGen} & \multicolumn{2}{c|}{--}  & \multicolumn{1}{c|}{OmniTok}  & \multicolumn{1}{c}{LlamaGen}\\\midrule
 
\(\boldsymbol{\alpha}\) & \textbf{FID50k$\downarrow$} & \textbf{IS$\uparrow$} &  \textbf{FID50k$\downarrow$} & \textbf{IS$\uparrow$} & \textbf{FID10k$\downarrow$} & \textbf{IS$\uparrow$} & \textbf{FVD$\downarrow$} & \textbf{FVD$\downarrow$} \\ 
\midrule
0.0 & 46.86          & 23.40          &    25.30          &  48.12          & 26.53           & 10.69          & 558.19           & 529.5             \\
0.1 & 36.03          & 30.56          &    20.65          &  54.98          & \textbf{20.78}  & \textbf{11.87} & 327.61           & 502.7             \\
0.2 & \textbf{32.47} & 34.87          & \textbf{19.83}    &  59.49          & 22.26           & 11.62          & \textbf{270.15}    & \textbf{476.9}  \\
0.3 & 33.57          & \textbf{35.42} & 21.03             &  \textbf{63.17} & 23.23           & 11.61          & 316.30             & 515.7           \\
\bottomrule
\end{tabular}
}
\caption{\textbf{Random Token Injection Analysis across dataset and architecture:} Effect of varying random token injection \(\alpha\) after 410k training steps, \rev{on class-to-image} ($\text{ImageNet}_{train}$, $\text{CIFAR-10}_{val}$), \rev{class-to-video} ($\text{UCF-101}_{test}$), and \rev{image-to-video} ($\text{NuScenes}_{train+val}$). \rev{$\alpha=0$ trains without injection and samples without correction.} Best results are highlighted in bold.}
\label{tab:alpha_ablation_32}
\end{table}

\subsection{Token Fixing}\label{sec:token_fixing}
\rev{\autoref{tab:alpha_ablation_32}, \autoref{tab:partiprompt} and \autoref{tab:kernel} are controlled: only the token injection or the sampler changes. The comparisons with other methods (\autoref{sec:c2i}, \autoref{sec:t2i}) also differ in data, tokenizer and training recipe.}

\paragraph{Random Token Injection $\alpha$.}
We analyze the impact of varying our main hyper-parameter $\alpha \in \{0, 0.1, 0.2, 0.3\}$ during training, as presented in~\autoref{tab:alpha_ablation_32} and ~\autoref{tab:partiprompt}. This parameter plays two key roles in the network's behavior. During training, it controls the amount of randomly injected tokens in the context, influencing the model’s ability to handle noisy inputs. When $\alpha > 0$, the model learns to be robust against perturbations. During sampling, setting $\alpha > 0$ enables the model to detect erroneous samples from previous iterations and refine predictions by better estimating the distribution over the input tokens. \rev{Any $\alpha>0$ improves over $\alpha=0$ on every dataset and metric. $\alpha=0.2$ gives the best ImageNet FID at every number of steps (\autoref{tab:ablation_step_and_alpha}) and the best FVD on both video datasets, but ImageNet IS keeps improving up to $\alpha=0.3$ and CIFAR-10 peaks at $\alpha=0.1$ with 32 steps; a value between $0.1$ and $0.2$ is a robust default. On video, the absolute FVD is bounded by the open-weight tokenizer (\autoref{sec:video}): these experiments show that token fixing transfers across modalities, not that it competes with the best video generators.}

\begin{figure}
\centering
\begin{subfigure}{0.49\linewidth}
    \centering
    \includegraphics[width=\linewidth]{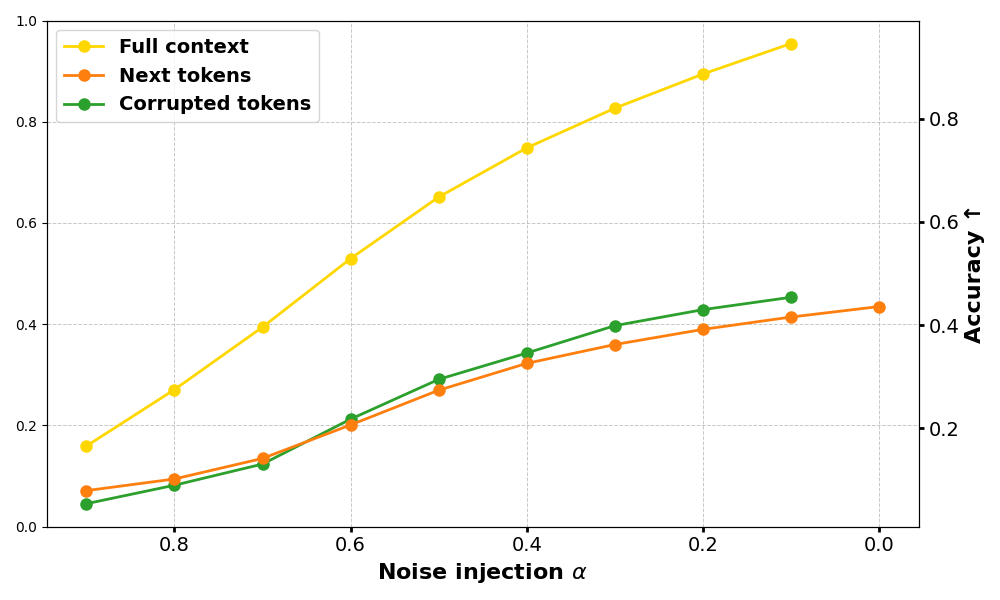}
    \subcaption{\textbf{Prediction Accuracy and rFID using  only 37\% tokens from a real image in context.} Each available token is then replace by a random tokens based on $\alpha$. Fixing corrupted tokens is as challenging as predicting next tokens.}
    \label{tab:resample_ablation}
\end{subfigure}
\hfill
\begin{subfigure}{0.49\linewidth}
    \centering
    \includegraphics[width=\linewidth]{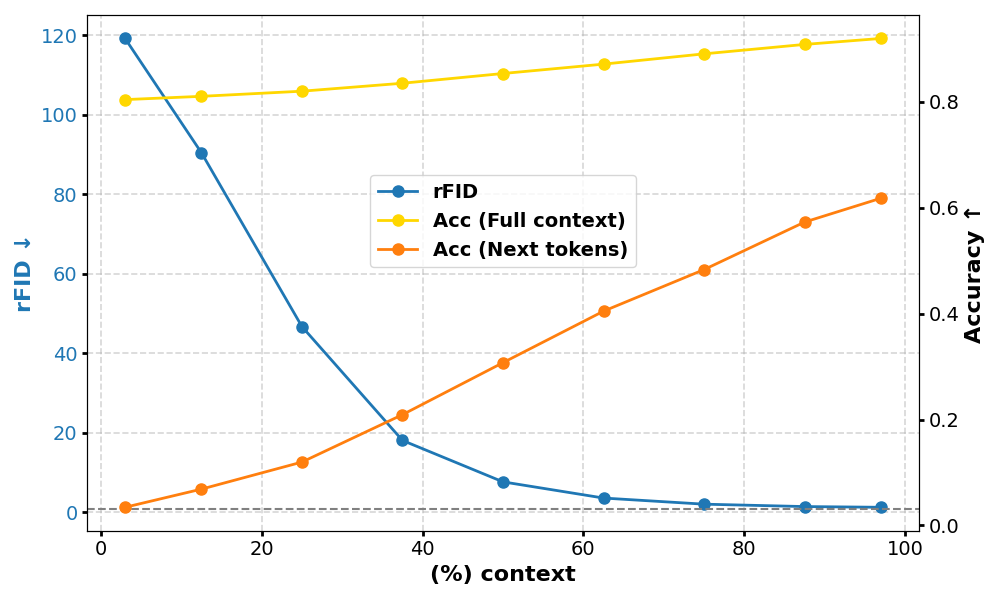}
    \subcaption{\textbf{Prediction Accuracy and rFID using constant $\alpha=0.2$.} We track the accuracy of one-step reconstructions as the proportion of the context increases. Most errors occur in the early stages, when only a few tokens are available in the context.}
    \label{fig:fid_acc_graph}
\end{subfigure}

\caption{\textbf{One-step reconstruction on ImageNet 256$\times$256 validation set.} Evaluation of model accuracy and fidelity under corrupted token contexts.}
\vspace{-0.5cm}
\label{fig:acc}
\end{figure}

\paragraph{Model Accuracy.} 
We evaluate in~\autoref{fig:acc} whether our model can accurately correct corrupted tokens and predict the next tokens among the $16{,}384$ codebook entries. In~\autoref{tab:resample_ablation}, we keep only 37\% tokens in the context, in which $n\times \alpha$ random tokens are injected. 
We then perform a single prediction step and measure whether the true label is among the top 1\% of predicted probabilities, reporting the metrics as Top-1\% accuracy. Specifically, we compute: 
(i) $\text{Acc}_{\text{next}}$, for predicting the next tokens; 
(ii) $\text{Acc}_{\text{full\_context}}$, for all tokens in the context; and 
(iii) $\text{Acc}_{\text{corrupted\_tokens}}$, computed only over the perturbed tokens in the context.
Our best model with $\alpha=0.2$ shows strong performance even under high perturbation. Moreover, it shows that the model handles unchanged tokens well, while fixing corrupted ones is as challenging as predicting next tokens, see discrepancy between green and yellow curve. In~\autoref{fig:fid_acc_graph}, most errors occur early, while later predictions improve, highlighting the need for a correction mechanism.

\begin{figure}[ht]
    \centering
    \begin{minipage}{0.49\linewidth}
        \centering
        \includegraphics[width=\linewidth]{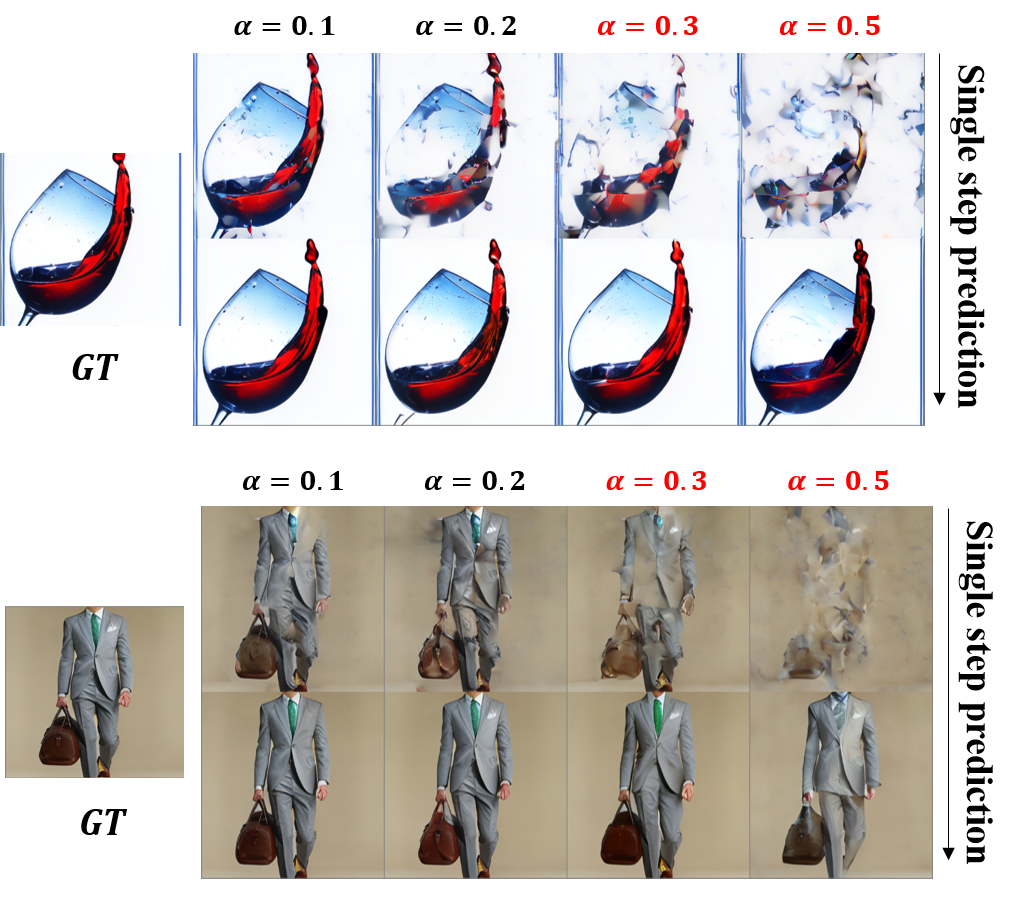}
        \caption{\textbf{Random token correction:} Top rows: random token injected at different values of $\alpha$, with no other tokens masked. Bottom rows: the model's prediction in a single step. The figure illustrates the model's ability to accurately correct tokens even when $\alpha$ exceeds the training range (i.e., $\textcolor{red}{\alpha > 0.2}$).}
        \label{fig:noise_correction}
    \end{minipage}
    \hfill
    \begin{minipage}{0.49\linewidth}
        \centering
        \includegraphics[width=\linewidth]{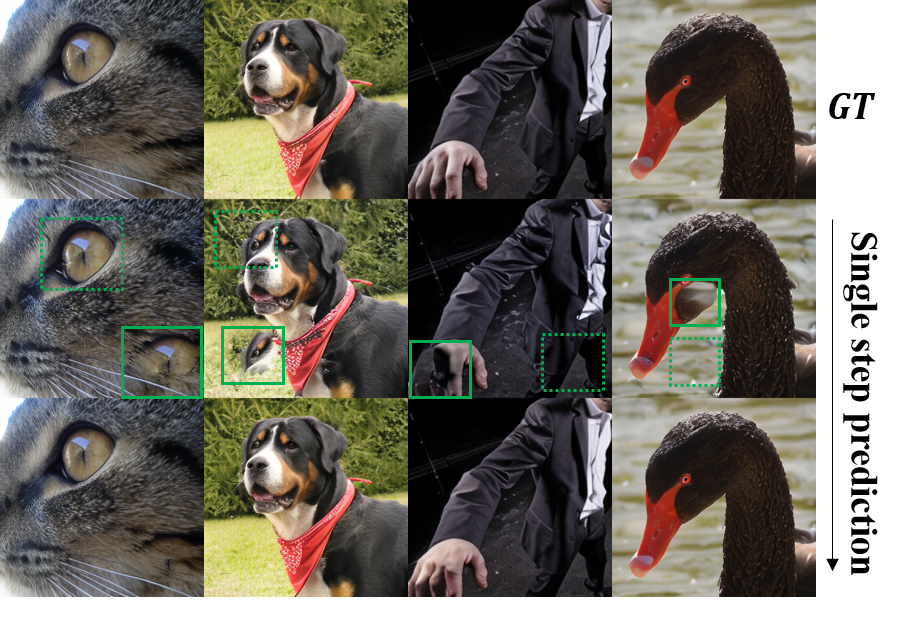}
        \caption{\textbf{Patch tokens correction:} Tokens from the green dotted square are copied and pasted into the solid green square, creating a corrupted image (second row). Only the tokens in the solid green square are altered, while all other tokens remain unchanged. \ours{} is able to correct these tokens in a single step (third row), demonstrating its ability to perform correction under out-of-distribution latent manipulations. Better visible when zoomed in.}
        \label{fig:copypast}
    \end{minipage}
    \vspace{-0.2cm}
\end{figure}
\paragraph{Self-Correction.} Here we keep all tokens in the context and demonstrate~\autoref{fig:noise_correction} the model’s ability to correct randomly injected tokens in the image. The model successfully recovers tokens in a single step even when $\alpha$ is set higher than the values used during training. We test the model’s ability to correct corrupted latent codes by copying a crop of GT tokens from one region and pasting them into a different location within the latent space. The modified latent code is passed through \ours{}, which fixes the token values in a single step. The model effectively corrects artifacts and local inconsistencies, such as missing fingers or duplicate eyes, as illustrated in~\autoref{fig:copypast}.

Visualization in~\autoref{fig:self_correction} showcases the model’s capacity to self-correct during sampling from scratch. For instance, it successfully prevents the generation of an extra mouth for the shark. On average, the model corrects 58 tokens per image over 32 sampling steps, corresponding to approximately 10\% of the final 576 tokens. \rev{A changed token is not necessarily a fixed error, since some changes are neutral; \autoref{tab:resample_ablation} therefore measures correction accuracy directly, on contexts where the corrupted tokens and their ground truth are known.}

\rev{
\subsection{Corruption Kernel and Correction Baselines}\label{sec:kernel}

\begin{table}[t]
\centering
\begin{tabular}{llcc}
\toprule
\textbf{Training corruption} & \textbf{Sampling} & \textbf{FID50k$\downarrow$} & \textbf{IS$\uparrow$} \\ \midrule
None ($\alpha=0$)                    & standard            & 25.30          & 48.12          \\
None ($\alpha=0$)                    & + ReMDM remasking   & 25.82          & \textbf{67.10} \\ \midrule
Uniform codebook ($\alpha=0.2$)      & with correction     & 23.93          & 50.90          \\
Model's own samples ($\alpha=0.2$)   & with correction     & 24.62          & 62.61          \\
In-image swap ($\alpha=0.2$, ours)   & with correction     & \textbf{19.83} & 59.49          \\
\bottomrule
\end{tabular}
\caption{\textbf{Corruption kernel and inference-time remasking on ImageNet 256$\times$256.} \ours{}-Base trained for 410k iterations and sampled in 16 steps without classifier-free guidance. Top: the backbone trained without injection ($\alpha=0$, the Halton-MaskGIT setting), sampled with the standard sampler or with ReMDM remasking~\citep{wang2025remasking}. Bottom: three corruption kernels at the same injection rate $\alpha=0.2$, all sampled with token correction.}
\label{tab:kernel}
\end{table}

\paragraph{Corruption kernel.}
\autoref{tab:kernel} compares three corruptions at $\alpha=0.2$: entries drawn uniformly from the codebook, as in BERT or discrete diffusion~\citep{devlin2019bert, austin2021structured}; tokens sampled by the model itself, which mimic its own errors; and the in-image swap of~\autoref{eq:random_injection}. All improve over $\alpha=0$, but the in-image swap leads by 4 to 5 FID points. Uniform codes are often out of distribution, hence easy to detect; self-sampled tokens raise IS but not FID, cost an extra forward pass per training step and are biased toward the current mode of the model. In-image swaps stay plausible and only break spatial consistency, which matches the structural errors that correction targets (see~\autoref{sec:mutual_info} for the intuition).

\paragraph{Inference-time remasking.}
ReMDM~\citep{wang2025remasking} remasks generated tokens at inference, without retraining. On the same backbone, tokenizer and steps, it raises IS at the cost of diversity, and FID does not improve (25.82 vs.\ 25.30), whereas \ours{} gains 5.5 FID points: a model never trained on corrupted contexts cannot reliably tell which of its tokens are wrong. Predictor-corrector diffusion~\citep{lezama2023discrete} and Token-Critic~\citep{lezama2022tokencritic} need a separately trained network and are not reimplemented; the Token-Critic numbers in~\autoref{tab:imagenet256} come from the original paper.
}

\subsection{Class to Image comparison}\label{sec:c2i}
We evaluate across different model sizes to assess their impact on performance, as shown in~\autoref{tab:ablation_model} \rev{(\autoref{sec:design_choices})}. Our model demonstrates faster convergence compared to Diffusion~\citep{Peebles2023scalable}, Flow Matching~\citep{ma2024sit}, and auto-regressive approaches~\citep{sun2024autoregressive} without the need of representation alignment~\citep{yu_representation_2025}. Using only 16 steps, we outperform them consistently, given comparable training steps and parameter counts.

We compare our method against SOTA in~\autoref{tab:imagenet256}. We use LlamaGen to encode images in \( 24 \times 24 \) visual tokens, producing an initial resolution of 384, which we downsampled to 256 following~\cite{sun2024autoregressive}. We train our model with $\alpha=0.2$, 32 steps, the Halton sequence, and the arccos scheduling. The full ablation is available in~\autoref{sec:design_choices}. When we prevent our model from correcting during sampling, we achieve an FID of 3.36, 0.38 better than the comparable Halton-MaskGIT baseline. Random token injection acts as data augmentation, enhancing robustness. Self-correction improves the FID by 0.87, reaching 2.49. Our approach is among the fastest sampling methods, as we require only 32 steps. Compared to VAR~\citep{tian2024visual}, we better cover the diversity of the real data (Recall). Additional results for UCF-101 and NuScenes are available in~\autoref{sec:video}. \rev{Only Halton-MaskGIT shares the architecture and tokenizer of \ours{}, so~\autoref{tab:imagenet256} positions \ours{} among discrete generators rather than isolating token fixing.}

\rev{\paragraph{Inference cost.} Correction adds no forward pass. \autoref{sec:speed} reports measured throughput for every model size on an A100 (e.g., 21.9 img/s at 8 steps and 6.2 at 32 steps for \ours{}-XLarge on $16\times16$ tokens), and \autoref{tab:transformer_config} the GFLOPs per forward pass. Since context tokens can change, correction rules out KV-caching.}

We present qualitative results in~\autoref{fig:self_correction}, ~\autoref{fig:teaser_imagenet}, and~\autoref{sec:qual_res}. \ours{} shows strong visual fidelity, effective error correction, and high diversity, highlighting the effectiveness of our approach.

\begin{table}
\centering
\resizebox{\textwidth}{!}{%
\begin{tabular}{l|lccc|cccc}
\toprule
\textbf{Type} & \textbf{Model}  & \#\textbf{Para.} & \textbf{Step} & \textbf{Training} & \textbf{FID}$\downarrow$ & \textbf{IS}$\uparrow$ & \textbf{Prec.}$\uparrow$ & \textbf{Rec.}$\uparrow$  \\ \midrule
\multirow{1}{*}{Conti.} 
& LDM-4~\citep{rombach2022ldm}               & 400M  & 250 & 0.2M & 3.60  & 247.7  & $-$  & $-$     \\
& DiT-XL/2~\citep{Peebles2023scalable}       & 675M  & 250 & 7.0M & 2.27  & \textbf{278.2}  & \textbf{0.83} & 0.57    \\
& SiT-XL/2~\citep{ma2024sit}                 & 675M  & 250 & 7.0M & \textbf{2.06}  & 270.3  & 0.82 & \textbf{0.59}    \\ \midrule
\multirow{1}{*}{AR}  
& Open-MAGVIT2-L~\cite{luo2024open}          & 804M  & 256 & 2.0M & \textbf{2.51} & 271.7 &  \textbf{0.84} & 0.54     \\
& LlamaGen-XL~\citep{sun2024autoregressive}  & 775M  & 576 & 1.6M & 2.62 & 244.1 & 0.80 & \textbf{0.57}     \\
& ZipAR~\citep{he2025zipar}                  & 775M  & 331 &  -   & 3.67 &  -    &  -   &  -                 \\
& VAR~\citep{tian2024visual}                 & 600M  & 10  & 2.0M & 2.57 & \textbf{302.6} & 0.83 & 0.56      \\ \midrule
\multirow{1}{*}{MIM} 
& MaskGIT~\citep{chang2022maskgit}           & 227M  & 12 & 1.6M & 6.18  & 182.1 & 0.80  & 0.51     \\
& TokenCritics~\citep{lezama2022tokencritic} & 454M  & 36 & 3.2M & 4.69  & 174.5 & 0.76  & 0.53     \\ 
& NAR~\citep{He_2025_ICCV}                   & 816M  & 31 &  -   & 2.70  & 277.5 & 0.81  & 0.58  \\
& Halton MaskGIT~\citep{besnier2025halton}   & 705M  & 32 & 2.0M & 3.74  & \textbf{279.5} & 0.81  & 0.60     \\ \midrule
\multirow{1}{*}{Ours}  
& \textbf{\ours{}-XLarge} - \textit{No cfg}  & 693M  & 32 & 2.0M & 6.06 & 145.9 &  0.75   &   0.65         \\
& \textbf{\ours{}-XLarge} - \textit{No Correction} & 693M  & 32 & 2.0M & 3.36 & 246.3 &  0.80  & 0.61           \\
& \textbf{\ours{}-XLarge}                   & 693M  & 32 & 2.0M & \textbf{2.49} & 252.5 &  \textbf{0.83}  & \textbf{0.63}       \\
\bottomrule
\end{tabular}
}
\vspace{-0.2cm}
\caption{\textbf{Class-conditional ImageNet 256$\times$256 benchmark.} We only include models with sizes below 1B. \rev{Baseline numbers are taken from the original papers, with their own tokenizers and training budgets; only Halton-MaskGIT shares the architecture and tokenizer of \ours{}. \textit{No Correction}: the same \ours{} model sampled without updating the context tokens.}}
\vspace{-0.5cm}
\label{tab:imagenet256}
\end{table}

\subsection{Text-to-Image Generation}\label{sec:t2i}

We train \ours{} for text-to-image generation using T5-XL~\citep{chung2024scaling} as the text encoder with standard cross-attention for conditioning. We evaluate our model on the PartiPrompts benchmark~\citep{yu2022scaling}. %

To study the effect of the noise injection parameter $\alpha$, we train \ours{}-Large on a mixture of CC12M~\citep{changpinyo2021cc12m} and synthetic images from JourneyDB~\citep{sun2023journeydb}, resulting in approximately 7M image-text pairs. As shown in \autoref{tab:partiprompt} \rev{(\autoref{sec:design_choices})}, increasing $\alpha$ \rev{from 0 to 0.2} improves generation quality across all evaluation metrics, matching the trend observed in other modalities. 

For our final model, we further expand the training data with the recently released GPIC~\citep{chandrasegaran2026gpic} and Fine-T2I~\citep{ma2026fine} datasets, resulting in a total of 106 million training samples. We first train at $256 \times 256$ resolution for 500k iterations with a batch size of 1024, followed by an additional 100k iterations at $512 \times 512$ and batch size 128. During the second stage, we consider two training variants: one with reward-conditioned post-training ($\dagger$), following~\citep{dufour2024dont,dufour2025miro}, and one without. \rev{The post-training uses only three rewards (CLIP score~\citep{radford2021learning}, PickScore and ImageReward), against seven for Miro.} Unlike RLHF approaches such as DPO or GRPO, reward conditioning does not require preference optimization. We also scale the model to the \ours{}-XLarge architecture. We compare against recent state-of-the-art methods, including PixArt-$\alpha$~\citep{chen2024pixartalpha}, SD3~\citep{esser2024scaling}, and FLUX~\citep{flux2024}. \rev{In \autoref{tab:sotatxt2img}, each variant is ranked against the baselines only. With reward conditioning, \ours{}-XLarge reaches an average rank of 3.25, the best below 1B parameters, in 32 steps; its closest reference is Miro~\citep{dufour2025miro} (5.50), which also uses multi-reward conditioning, while FLUX and SD3 rely on preference fine-tuning. Without post-training, it ranks 6.25, behind PixArt-$\alpha$, PixArt-$\Sigma$ and Miro.}

\rev{\paragraph{Compositional evaluation.} Global preference metrics only partially capture the structural errors that token fixing targets. On GenEval~\citep{ghosh2023geneval} (\autoref{tab:geneval}), which checks object count, color, position and attribute binding with a detector, \ours{}-XLarge with reward conditioning scores 0.65 overall, above SDXL and PixArt-$\alpha$ and close to the 20 times larger FLUX-dev (0.67). Miro reaches 0.75, but with seven rewards against three for \ours{}. Compared with Meissonic~\citep{bai2025meissonic}, the closest masked-transformer baseline, \ours{} gains 0.11 overall and 0.39 on position, although the two models differ in training data, size and training recipe. The largest gain over FLUX-dev is on position (0.49 vs.\ 0.20).}

\begin{figure}[p]
    \centering
    \begin{subfigure}{\textwidth}
        \centering
        \includegraphics[width=0.16\textwidth]{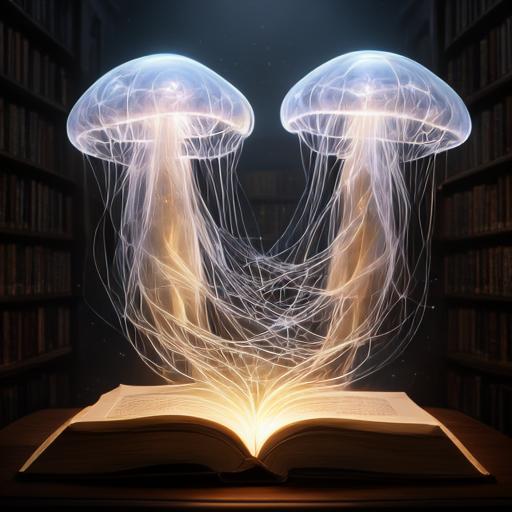}\hspace{-0.2cm}
        \includegraphics[width=0.16\textwidth]{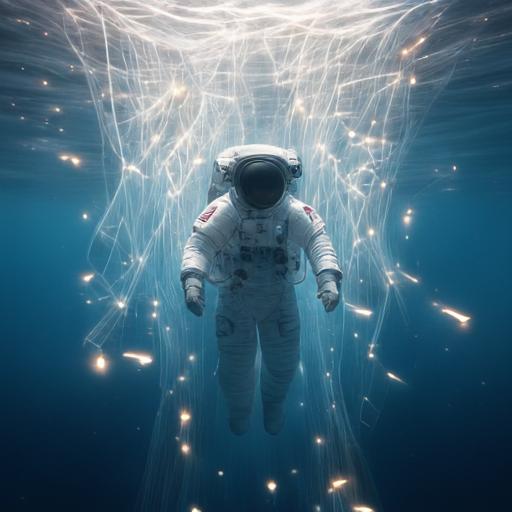}\hspace{-0.2cm}
        \includegraphics[width=0.16\textwidth]{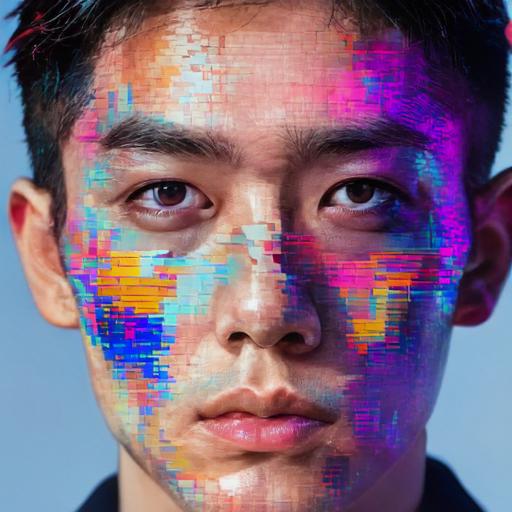}\hspace{-0.2cm}
        \includegraphics[width=0.16\textwidth]{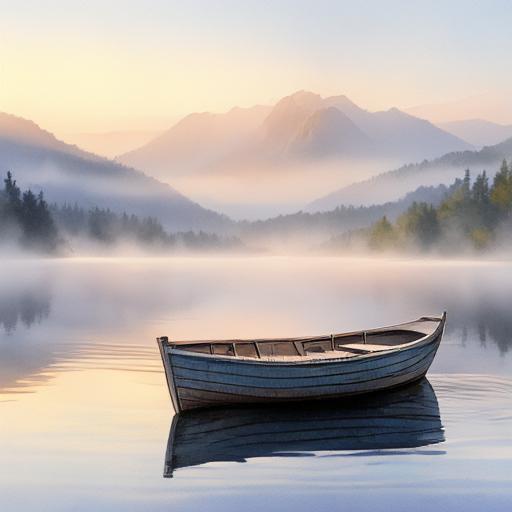}\hspace{-0.2cm}
        \includegraphics[width=0.16\textwidth]{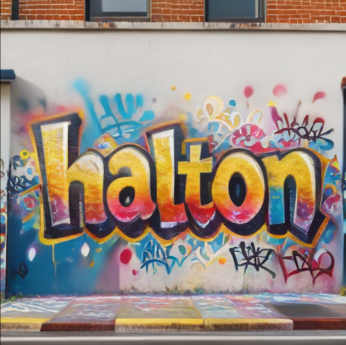}\hspace{-0.2cm}
        \includegraphics[width=0.16\textwidth]{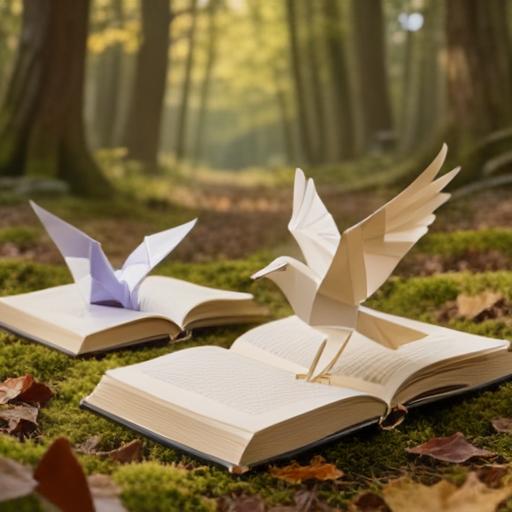}\\
        \includegraphics[width=0.16\textwidth]{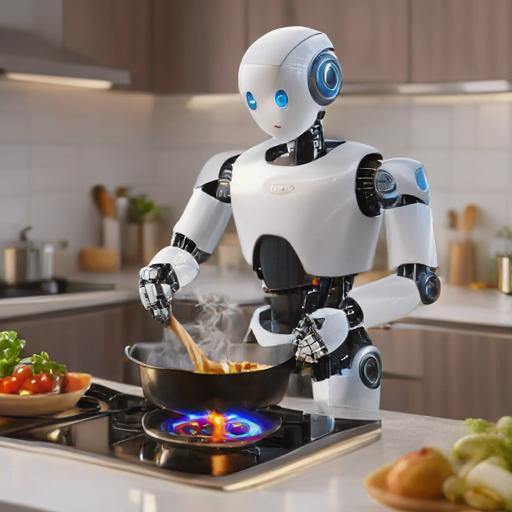}\hspace{-0.2cm}
        \includegraphics[width=0.16\textwidth]{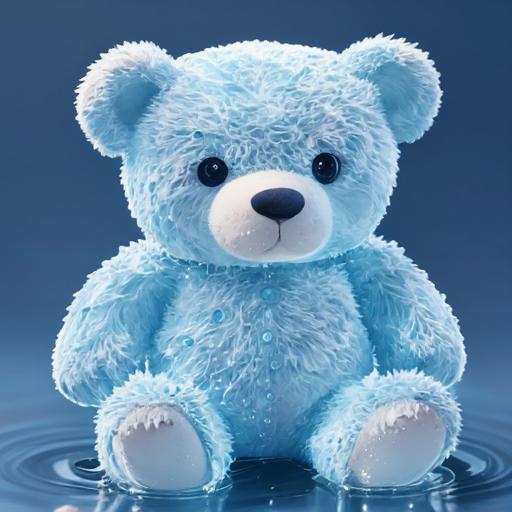}\hspace{-0.2cm}
        \includegraphics[width=0.16\textwidth]{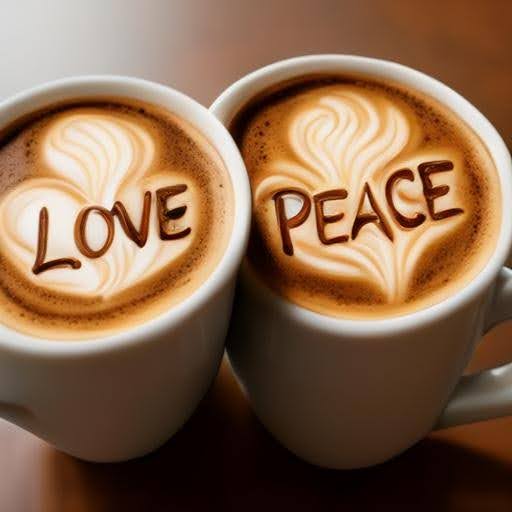}\hspace{-0.2cm}
        \includegraphics[width=0.16\textwidth]{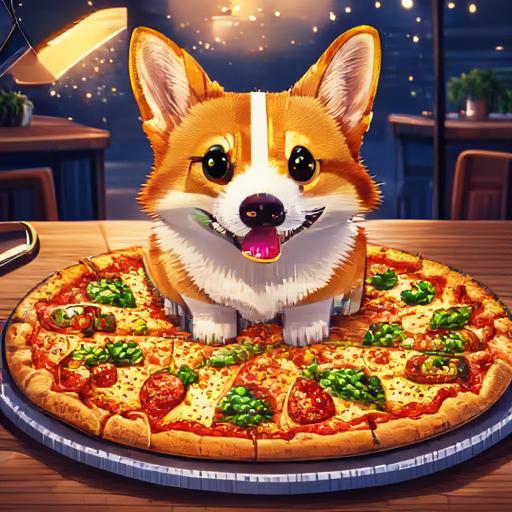}\hspace{-0.2cm}
        \includegraphics[width=0.16\textwidth]{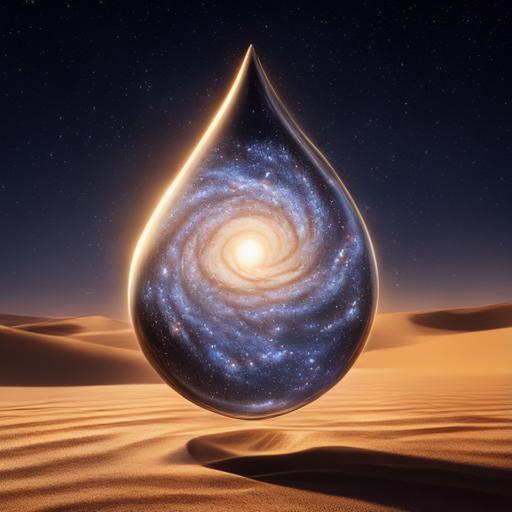}\hspace{-0.2cm}
        \includegraphics[width=0.16\textwidth]{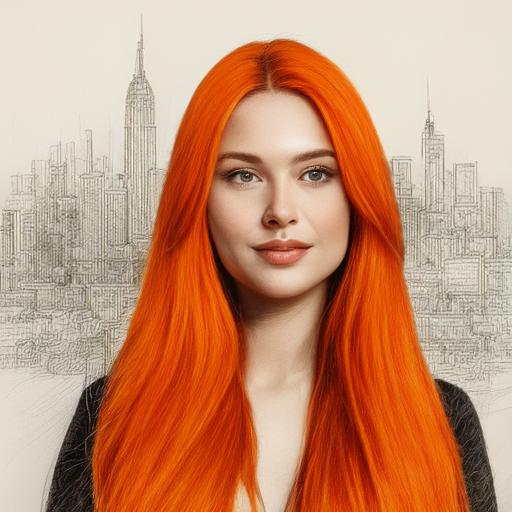}
        \caption{Text-to-image generation samples.}
        \label{fig:teaser_t2i}
    \end{subfigure}

    \vspace{0.5em}

    \begin{subfigure}{\textwidth}
        \centering
        \includegraphics[width=0.955\textwidth]{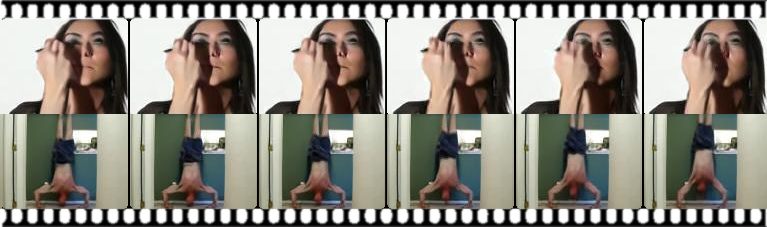}
        \caption{Video generation snapshot.}
        \label{fig:teaser_video}
    \end{subfigure}

    \vspace{0.5em}

    \begin{subfigure}{\textwidth}
        \centering
        \includegraphics[width=0.16\textwidth]{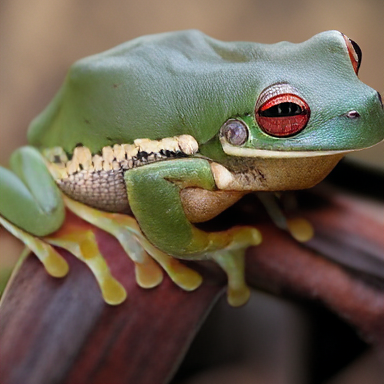}\hspace{-0.2cm}
        \includegraphics[width=0.16\textwidth]{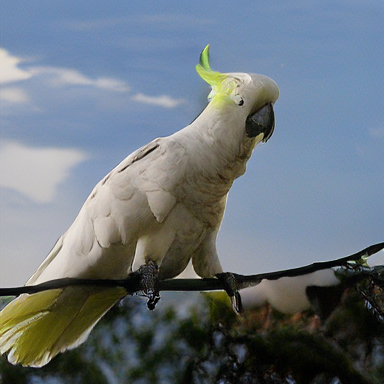}\hspace{-0.2cm}
        \includegraphics[width=0.16\textwidth]{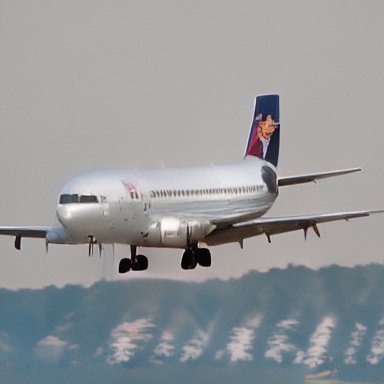}\hspace{-0.2cm}
        \includegraphics[width=0.16\textwidth]{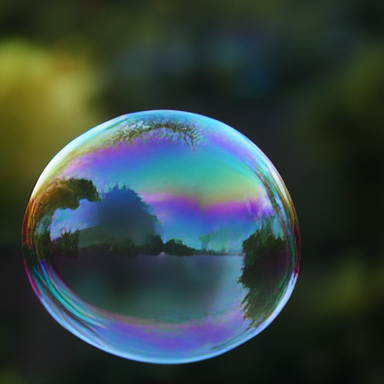}\hspace{-0.2cm}
        \includegraphics[width=0.16\textwidth]{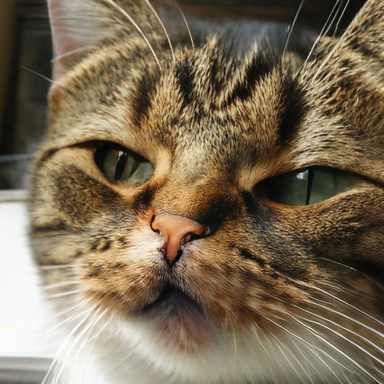}\hspace{-0.2cm}
        \includegraphics[width=0.16\textwidth]{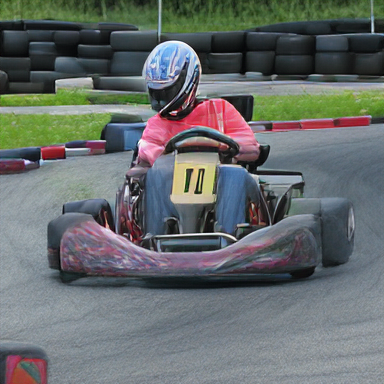}\\

        \includegraphics[width=0.16\textwidth]{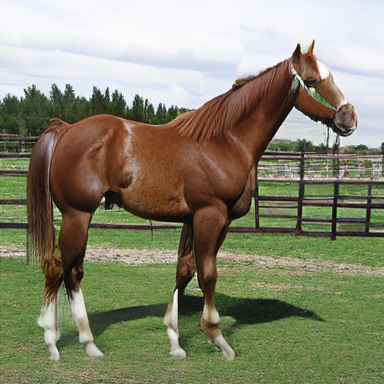}\hspace{-0.2cm}
        \includegraphics[width=0.16\textwidth]{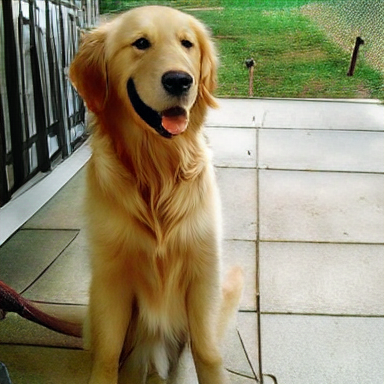}\hspace{-0.2cm}
        \includegraphics[width=0.16\textwidth]{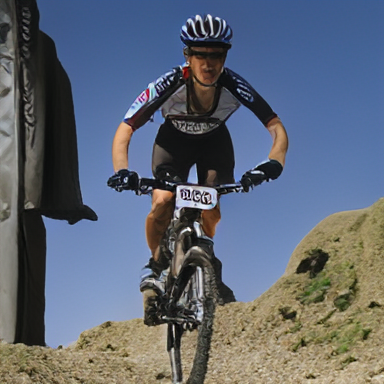}\hspace{-0.2cm}
        \includegraphics[width=0.16\textwidth]{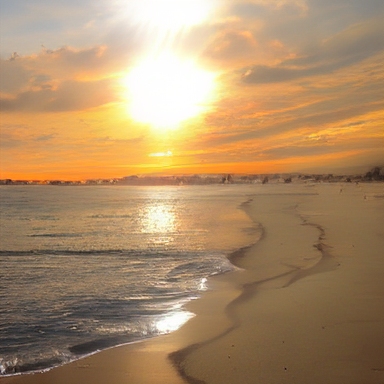}\hspace{-0.2cm}
        \includegraphics[width=0.16\textwidth]{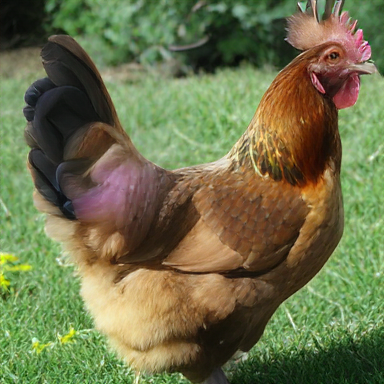}\hspace{-0.2cm}
        \includegraphics[width=0.16\textwidth]{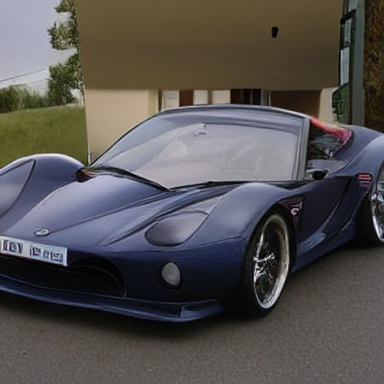}
        \caption{ImageNet 256$\times$256 samples.}
        \label{fig:teaser_imagenet}
    \end{subfigure}
    
    \caption{\textbf{Qualitative results.}
    (\subref{fig:teaser_t2i}) Text-to-image generation results showcasing high image quality, accurate text rendering, and diverse compositions across prompts.
    (\subref{fig:teaser_video}) Consecutive frames from generated videos, demonstrating strong temporal consistency while preserving object appearance and scene dynamics.
    (\subref{fig:teaser_imagenet}) ImageNet samples generated by our largest model, illustrating high fidelity to the data distribution and photorealistic image synthesis.}    
\end{figure}

Qualitative examples are shown in \autoref{fig:teaser_t2i} on PartiPrompts, Prism, and custom prompts. The model produces high-quality images with strong text alignment, accurate text rendering, and diverse compositions.

\begin{table}
\centering
\resizebox{\columnwidth}{!}{
\begin{tabular}{lccccccc}
\toprule
\textbf{Methods} & \textbf{\#Para.} & \textbf{Avg Rank$\downarrow$} & \textbf{Aesthetic$\uparrow$} & \textbf{ImageReward$\uparrow$} & \textbf{HPSv2$\uparrow$} & \textbf{PickScore$\uparrow$} \\
\midrule
FLUX-dev~\citep{flux2024}                        & 12B  & \textbf{1.25} & \textbf{6.56} & 1.19 & \textbf{0.30} & \textbf{0.229} \\
Sana-1.6B~\citep{xie2024sana}                    & 1.6B & 2.00          & 6.36          & \textbf{1.23} & \textbf{0.30} & 0.228 \\
SD3-medium~\citep{esser2024scaling}              & 2.0B & 4.25          & 6.18          & 1.15          & \textbf{0.30} & 0.225 \\
PixArt-$\alpha$~\citep{chen2024pixartalpha}      & 0.6B & 4.50          & 6.47          & 0.97          & 0.29           & 0.226 \\
PixArt-$\Sigma$~\citep{chen2024pixart}           & 0.6B & 4.75          & 6.44          & 1.02          & 0.29           & 0.225 \\
SDXL~\citep{podell2024sdxl}                      & 2.6B & 8.50          & 5.94          & 0.46          & 0.25           & 0.220 \\
SD v2.1~\citep{rombach2022ldm}                   & 0.9B & 9.00          & 5.81          & 0.38          & 0.26           & 0.215 \\
CAD~\citep{dufour2024dont}                       & 0.4B & 9.25          & 5.56          & 0.69          & 0.26           & 0.214 \\
SD v1.5~\citep{rombach2022ldm}                   & 0.9B & 10.50         & 5.68          & 0.24          & 0.25           & 0.213 \\
Miro~\citep{dufour2025miro}                       & 0.4B & 5.50          & 6.28          & 1.11          & 0.29           & 0.220 \\\midrule
\textbf{\ours{}-XLarge ($\alpha{=}0.2$) (ours)}        & 0.6B & \rev{6.25}         & 6.25          & 1.14          & 0.28          & 0.219 \\
\textbf{\ours{}-XLarge ($\alpha{=}0.2$) (ours)}$\dagger$        & 0.6B & 3.25          & 6.23          & 1.16          & \textbf{0.30}          & 0.227 \\

\bottomrule
\end{tabular}
}
\caption{\textbf{\texttt{PartiPrompt} benchmark.} \rev{Four automatic preference metrics and their average rank. Each \ours{} variant is ranked against the baselines only; ties share the same rank. \ours{} uses 32 sampling steps.} $\dagger$: finetuned model with reward conditioning.}
\label{tab:sotatxt2img}
\end{table}

\rev{
\begin{table}[t]
\centering
\resizebox{\columnwidth}{!}{
\begin{tabular}{lcccccccc}
\toprule
\textbf{Methods} & \textbf{\#Para.} & \textbf{Overall$\uparrow$} & \textbf{Single obj.$\uparrow$} & \textbf{Two obj.$\uparrow$} & \textbf{Counting$\uparrow$} & \textbf{Colors$\uparrow$} & \textbf{Position$\uparrow$} & \textbf{Color attr.$\uparrow$} \\
\midrule
PixArt-$\alpha$~\citep{chen2024pixartalpha} & 0.6B & 0.48 & 0.98 & 0.50 & 0.44 & 0.80 & 0.08 & 0.07 \\
SDXL~\citep{podell2024sdxl}                 & 2.6B & 0.55 & 0.98 & 0.74 & 0.39 & 0.85 & 0.15 & 0.23 \\
FLUX-dev~\citep{flux2024}                   & 12B  & 0.67 & \textbf{0.99} & \textbf{0.81} & \textbf{0.79} & 0.74 & 0.20 & 0.47 \\
Miro$^\ddagger$~\citep{dufour2025miro}       & 0.4B & \textbf{0.75} & 0.98 & 0.79 & 0.71 & 0.85 & \textbf{0.58} & \textbf{0.58} \\
Meissonic~\citep{bai2025meissonic}          & 1.0B & 0.54 & \textbf{0.99} & 0.66 & 0.42 & \textbf{0.86} & 0.10 & 0.22 \\ \midrule
\textbf{\ours{}-XLarge ($\alpha{=}0.2$) (ours)}$\dagger$ & 0.6B & 0.65 & 0.94 & 0.74 & 0.63 & 0.81 & 0.49 & 0.29 \\
\bottomrule
\end{tabular}
}
\caption{\textbf{GenEval benchmark.} \rev{Fraction of generated images judged correct by the GenEval object detector for each category of prompts, and their average (Overall). $\dagger$: finetuned model with reward conditioning. $\ddagger$: Miro sampled with its custom reward target (aesthetic target set to 0.625).}}
\label{tab:geneval}
\end{table}
}

\section{Conclusion}
After Theoros built his new image synthesis framework, Eritos asks him if his idea worked.

\dialogue{T} “Absolutely! We introduced \ours{} a bidirectional image generation framework with token correction, addressing key limitations of existing multi-token prediction methods. During training, we inject \textit{random tokens} to enhance the model's robustness to errors. The model learns to detect inconsistencies in the context, fix them, and predict the next tokens of the sequence. When sampling, rather than simply generating new tokens at each step, our method actively refines the context, progressively fixing errors introduced in earlier stages. Our method achieves consistent improvements in image synthesis on ImageNet, CIFAR-10 or GPIC, as well as in video generation on UCF-101 and NuScenes. \ours{} method provides a better balance of speed and quality for visual synthesis.

\dialogue{E} “Amazing, I guess this means I’ll be seeing more chicken images from you soon with this?"

\dialogue{T} “You bet. Get ready for the most photorealistic generated chicken ever! (see~\autoref{fig:chicken})"

\rev{
\section*{Broader Impact Statement}
Like other image and video generators, \ours{} can be used to produce misleading or harmful synthetic media. The text-to-image models are trained on GPIC~\citep{chandrasegaran2026gpic}, a permissively licensed corpus, on Fine-T2I~\citep{ma2026fine}, on the web-collected CC12M~\citep{changpinyo2021cc12m}, and on synthetic images from JourneyDB~\citep{sun2023journeydb}. Web-scale image-text data carries social biases that the models can reproduce, and the licensing of individual web images is not always clear. Generated content should therefore be screened with existing safety classifiers before any deployment.
}

\bibliography{main}
\bibliographystyle{tmlr}

\appendix
\clearpage

\appendix

\section{On Mutual Information minimization}\label{sec:mutual_info}

\rev{This section gives the intuition behind the in-image corruption kernel; it is not a formal derivation, and the mutual information is not measured. The empirical support for the kernel is the ablation of~\autoref{sec:kernel}.}

With no loss of generality, let the visual latent be \(\mathcal{Z} = \{z_1, z_2, z_3, z_4\} \). We would like to model the conditional distribution \(P(z_3,z_4 \mid z_1,z_2)\). Because learning the full joint conditional \(P(z_3,z_4 \mid z_1,z_2)\) in one step can be difficult, a common approach is to learn the two separate conditionals
\(P(z_3 \mid z_1,z_2)\) and \(P(z_4 \mid z_1,z_2)\) and approximate \( P(z_3,z_4 \mid z_1,z_2) \approx P(z_3 \mid z_1,z_2)\,P(z_4 \mid z_1,z_2). \)

The approximation error introduced by this factorization is composed of the mutual information~\citep{cover1999elements} (MI) between \(z_3\) and \(z_4\) given the context \((z_1,z_2)\) plus a residual error $E$ due to the underfitting of the model or the (aleatoric) uncertainty in the data. In this paper, we only tackle MI which measures how much knowing \(z_3\) reduces our uncertainty about \(z_4\), and leave the error $E$ as a training optimization problem. To correct MI, we introduce a noising step and a denoising/fixing step (in parallel with the next tokens prediction, without extra cost).

The corruption process introduced in~\autoref{sec:Method} can be viewed as a stochastic mapping
\(
\widetilde{\mathcal Z}
=
\Phi_{\alpha}(\mathcal Z),
\)
where each token is independently preserved or replaced according to Eq.~(\ref{eq:random_injection}). 
We then train a model to invert/fix the noisy token by learning the clean distribution using a cross-entropy objective.
\[ P(\mathcal{Z} \mid \widetilde{\mathcal{Z}}) = P(z_1,z_2,z_3,z_4 \mid \tilde{z}_1,\tilde{z}_2,\tilde{z}_3,\tilde{z}_4),\] (This corresponds to Eq.~\ref{eq:l_context}.)

In practice, any \(\Phi_{\alpha}\) that strongly increases MI of the tokens would push the model to learn to decrease the MI (i.e., to fix errors due to token correlations). We use the mapping in Eq.~(5), which swaps random tokens within the same image, ensuring high MI between them since tokens from the same image are indeed correlated. Thus the factorized approximation \(P(z_3 \mid \tilde{z}_1,\tilde{z}_2)P(z_4 \mid \tilde{z}_1,\tilde{z}_2)\) plus the fixing step yields a better approximation to the true conditional joint distribution \(P(z_3,z_4 \mid z_1,z_2)\).

\section{Video Synthesis}\label{sec:video}
To test our method beyond image synthesis, we explore class-to-video on UCF101~\citep{ucf101} dataset and img-to-video on NuScenes~\citep{caesar2020nuscenes}. As demonstrated previously in Table~\ref{tab:alpha_ablation_32}, introducing self-correction substantially improves the quality of generated video samples, mirroring the results of our image synthesis experiments. 

In~\autoref{tab:ucf101}, we compare \ours{}-Large against state-of-the-art video generation models on UCF101. Despite using a smaller model (480M parameters) and fewer training steps (32), our approach achieves a competitive FVD of 242.16. Performance is limited by our reliance on the open-weight OmniTokenizer, which yields a higher rFVD (42) compared to closed-source tokenizers used by MAGVIT (rFVD 25) and MAGVIT2 (rFVD 8.62). This highlights that while our framework is efficient, generative quality remains constrained by tokenization quality and scale. These results demonstrate the potential of \ours{} but indicate that further improvements would require larger models or more advanced tokenizers. \rev{The purpose of the video experiments is controlled evidence that token fixing transfers across modalities: with the same backbone and tokenizer, FVD improves from 558 to 270 on UCF-101 and from 530 to 477 on NuScenes when $\alpha$ goes from 0 to 0.2 (\autoref{tab:alpha_ablation_32}).}

\begin{table}[ht]
    \centering
        \begin{tabular}{lcccc}
    \toprule
    \textbf{Model} &  \#\textbf{Para.}   & \textbf{Train (steps)}   &  \textbf{Steps}     & \textbf{FVD$\downarrow$}  \\ \midrule
     MAGVIT~\cite{yu2023magvitmaskedgenerativevideo}$\dagger$ & 306M    & -       &   12   &  76              \\
     MAGVIT2~\citep{yu2024magvit2}$\dagger$                    & 307M    & -       &   24   &  58              \\
     LARP~\citep{wang2025larp}$\dagger$                        & 632M    & -       &   1024 &  \textbf{57}     \\\midrule
     OmniTokenizer~\citep{wang2024omnitokenizer}      & 650M    & 4M      &   1280 &  191.14          \\
     OmniTokenizer~\citep{wang2024omnitokenizer}      & 227M    & 4M      &   1280 &  313.14          \\
     \textbf{\ours{}-Large (our)}                          & 480M    & 0.4M    &     32 &  242.16 \\
    \bottomrule
    \end{tabular}
    \caption{UCF101 results. $\dagger$ use closed source tokenizer.}
    \label{tab:ucf101}
\end{table}

In~\autoref{tab:Nuscenes}, we report results on the NuScenes dataset. Despite using a relatively small model (\ours{}-Large, 480M parameters) and limited training time (15 hours), our method achieves a competitive FVD of 290.5. Compared to much larger models, such as GenAD (2.7B parameters, FVD 184.0) or Vista (2.5B parameters, FVD 89.4), \ours{} demonstrates strong efficiency and highlights the potential for scaling to larger models and longer training to achieve a lower FVD. 

\begin{table}[ht]
    \centering
    \begin{tabular}{lccccc}
    \toprule
    \textbf{Model}  &  \#\textbf{Para.} & \textbf{Train (h)} & \textbf{Steps} & \textbf{FVD} \\ \midrule
    GenAD~\citep{zheng2024genad}$\ddagger$           & 2.7B  & 2 kh   & --   & 184.0 \\
    Vista~\citep{gao2024vista}$\ddagger$           & 2.5B  & 1.7 kh & 100  & 89.4 \\\midrule
    DriveDreamer~\citep{wang2024drivedreamer}$\dagger$      & 1.45B & 15 h  & --   & 452.0 \\
    WoVoGen~\citep{lu2024wovogen}           & -     & 15 h   & --   & 417.7 \\
    Drive-WM~\citep{wang2024driving}$\dagger$          & 1.45B & 15 h  & 50   & 122.7 \\
    \textbf{\ours{}-Large (ours)}     & 480M  & 15 h   & 32+8 & 290.5 \\ \bottomrule
    \end{tabular}
    \caption{NuScenes results. $\dagger$ use pre-trained weight from SD~\citep{rombach2022ldm}. $\ddagger$ Zero-shot FID.\label{tab:nuscenes}}
    \label{tab:Nuscenes}
\end{table}

\section{Design Choices}\label{sec:design_choices}

In the following section, we investigate \textcolor{teal}{random token injection $\alpha$} in \autoref{tab:ablation_step_and_alpha}, the number of \textcolor{blue}{prediction steps} in~\autoref{tab:ablation_step_and_alpha}, followed by the type of \textcolor{darkorange}{scheduling methods} in~\autoref{tab:scheduling_methods}, and \textcolor{olive}{token ordering} in~\autoref{tab:ablation_init}. Each factor influences efficiency and accuracy, as discussed below.

\paragraph{Number of Steps}
We investigate in~\autoref{tab:ablation_step_and_alpha} the effect of the total number of steps $\{8, 16, 24, 32\}$ to predict the full images. On ImageNet, increasing the number of steps improves performance up to $step=16$, beyond which the benefits plateau. On the other hand, increasing the number of steps to 24 leads to improved results on Cifar10, suggesting that the step count should be scaled proportionally to the total number of tokens to be predicted. Moreover, increasing $\alpha$ leads the model to correct more tokens during inference, but the number of corrected tokens does not necessarily correlate with FID. \rev{The count includes every token that changes, and some changes are neutral, so it is not the number of errors fixed; the accuracy on known corrupted tokens (\autoref{tab:resample_ablation}) is the direct measure of correction.} Interestingly, the model is more sensitive to $\alpha$ than to the number of decoding steps. For example, a model trained with 8 decoding steps corrects nearly the same number of tokens as a model trained with 32 steps.

\begin{table}[ht]
    \centering
    \begin{tabular}{cc|ccc|cc}\toprule
                             &                        &        \multicolumn{3}{c}{\textbf{ImageNet}}                                   &             \multicolumn{2}{c}{\textbf{Cifar10}}      \\\midrule
\textbf{Step}                  &  \textbf{\(\alpha\)}   &  \textbf{FID50k$\downarrow$} & \textbf{IS$\uparrow$} &  \#\textbf{Avg. Cor} & \textbf{FID10k$\downarrow$} & \textbf{IS$\uparrow$}  \\ 
\midrule
\multirow{4}{*}{8}             & 0.0                    &      42.84            &  29.17                       & 0.00  & 86.58                       &      7.57               \\ 
                             & 0.1                    &     38.70             &  33.38                         & 4.47  & 80.30	                      &      7.82               \\
                             & 0.2                    &     \textbf{34.93}    &  36.68                         & 15.91 & 67.14	                      &      8.43               \\
                             & 0.3                    &     34.98             &  \textbf{37.6}                 & 38.15 & \textbf{66.52}	          &      \textbf{8.64}      \\\midrule
\multirow{4}{*}{16}           & 0.0                    &      43.76             &  26.21                       & 0.00  & 39.30                       &      9.86               \\
                             & 0.1                    &     35.94             &  32.71                         & 7.22  & 26.88	                      &      11.19              \\
                             & 0.2                    &     \textbf{31.27}    &  37.79                         & 22.69 & \textbf{25.76}	          &      \textbf{11.06}     \\
                             & 0.3                    &     32.66             &  \textbf{39.09}                & 50.14 & 27.04	                      &      \textbf{11.06}     \\\midrule
\multirow{4}{*}{24}           & 0.0                    &      45.15             &  24.68                       & 0.0  & 30.50                       &      10.85              \\ 
                             & 0.1                    &     35.82             &  31.75                         & 8.77  & \textbf{20.66}	          &      \textbf{11.77}     \\
                             & 0.2                    &     \textbf{31.96}    &  35.77                         & 26.79 & 22.07	                      &      11.44              \\
                             & 0.3                    &     31.98             & \textbf{37.81}                 & 58.21 & 22.85	                      &      11.46     \\ \midrule
\multirow{4}{*}{32}          & 0.0                    &      46.86             &  23.40                        & 0.00  & 26.53                       &      10.69              \\
                             & 0.1                    &     36.03             &  30.56                         & 10.23 & \textbf{20.78}	          &      \textbf{11.87}     \\
                             & 0.2                    &     \textbf{32.47}    &  34.87                         & 30.47 & 22.26	                      &      11.62              \\ 
                             & 0.3                    &     33.57             & \textbf{35.42}                 & 65.41 & 23.23	                      &      11.61              \\ 
        \bottomrule
        \end{tabular}
    \caption{\textbf{ImageNet 256 / Cifar10:} Ablation on the \textcolor{teal}{random token injection $\alpha$}, the number \textcolor{blue}{prediction steps} and the number of corrected token during inference, without cfg. We show that enabling token fixing, i.e., $\alpha>0$, largely improves the metrics, while $16$ steps is a good trade-off between FID/IS and compute efficiency. }
    \label{tab:ablation_step_and_alpha}
\end{table}

\begin{table}

\centering
\begin{adjustbox}{width=0.8\linewidth}
\begin{tabular}{lccccc}
\toprule
\textbf{Model}                               &  \#\textbf{Para.}   &  \textbf{Training}  &  \textbf{Steps}     & \textbf{FID50k$\downarrow$} & \textbf{IS$\uparrow$}  \\ \midrule
DiT-S/2~\citep{Peebles2023scalable}          &  33M                &  400k               &  250	               &   68.40   &  -     \\
SiT-S/2~\citep{ma2024sit}                    &  33M                &  400k               &  250	               &   57.60   &  -     \\
Halton-MaskGIT-S~\citep{besnier2025halton}   &  69M                & 410k                & 32 & 38.49 & 32.80 \\
\textbf{\ours{}-Small (our)}                 &  50M  	           &  410k               &  \textbf{16}	               &   \textbf{31.27}    &  \textbf{37.79}    \\\midrule 

DiT-B/2~\citep{Peebles2023scalable}          &  130M               &  400k               &  250	               &    43.47   &  -     \\
SiT-B/2~\citep{ma2024sit}                    &  130M               &  400k               &  250	               &    33.50   &  -     \\
LlamaGen-B~\citep{sun2024autoregressive}     &  111M               &  530k               &  256	               &    33.44  & 37.53 \\
Halton-MaskGIT-B~\citep{besnier2025halton}   &  142M               &  410k                & 32 & 24.91 & 58.98 \\
\textbf{\ours{}-Base (our)}                  &  143M	           &  410k               &  \textbf{16}	               &   \textbf{19.83}    &  \textbf{59.49}    \\\midrule

DiT-L/2~\citep{Peebles2023scalable}          &  458M               &  400k               &  250	               &    23.30   &  -     \\
SiT-L/2~\citep{ma2024sit}                    &  458M               &  400k               &  250	               &    18.80   &  -     \\
LlamaGen-L~\citep{sun2024autoregressive}     &  343M               &  530k               &  256	               &    19.07   & 64.35   \\
Halton-MaskGIT-L~\citep{besnier2025halton}   &  480M               &  410k                & 32 & 14.31 & 84.32 \\

\textbf{\ours{}-Large (our)}                 &  480M	           &  410k               &  \textbf{16}	       &   \textbf{11.36} & \textbf{95.17} \\\midrule

DiT-XL/2~\citep{Peebles2023scalable}         &  675M               &  400k               &  250	               &    19.50   &  -     \\
SiT-XL/2~\citep{ma2024sit}                   &  675M               &  400k               &  250	               &    17.20   &  -     \\
LlamaGen-XL~\citep{sun2024autoregressive}    &  775M               &  530k               &  256	               &    18.04   &  69.88 \\
\textbf{\ours{}-XLarge (our)}                &  693M	           &  410k               &  \textbf{16}	       &   \textbf{9.25} & \textbf{103.60} \\\bottomrule
\end{tabular}
\end{adjustbox}
\vspace{-0.3cm}
\caption{\textbf{Scaling model size on class-conditional ImageNet 256$\times$256 benchmark.} Analysis of model sizes, without classifier-free guidance. Our method converges faster than other competitive methods.}
\vspace{-0.5cm}
\label{tab:ablation_model}
\end{table}

\begin{table}{
\centering
\resizebox{\columnwidth}{!}{
\begin{tabular}{lcccccc}
\toprule
\textbf{Methods} & \textbf{$\alpha$} & \textbf{Train Steps} & \textbf{Aesthetic} & \textbf{ImageReward} & \textbf{HPSv2} & \textbf{PickScore} \\\midrule
\ours{}-Large    & 0.0 & 500k & 5.34 & 0.17 & 0.23 & 0.210 \\
\ours{}-Large    & 0.1 & 500k & 5.60 & 0.39 & \textbf{0.25} & 0.215 \\
\ours{}-Large    & 0.2 & 500k & \textbf{5.71} & \textbf{0.53} & \textbf{0.25} & \textbf{0.217} \\
\bottomrule
\end{tabular}
}
\caption{\textbf{Effect of $\alpha$ parameters on text-to-image model.} \rev{\ours{}-Large trained for 500k iterations on about 7M image-text pairs with $\alpha \in \{0, 0.1, 0.2\}$, evaluated on the \texttt{PartiPrompt} benchmark with four automatic preference metrics (higher is better).}}

\label{tab:partiprompt}
}
\end{table}

\paragraph{Influence of Scheduling Strategy}
To analyze the effect of different scheduling methods we measure performance across various configurations $\{\text{square, arccos, linear, root, constant}\}$, and show the results in~\autoref{tab:scheduling_methods}. Similarly to~\cite{chang2023muse} finding, we find out that the arccos scheduling performs the best while concave scheduling performs worse.

\begin{table}[ht]
    \centering
    \begin{minipage}{0.48\textwidth}  %
        \centering
        \begin{tabular}{l|cc}
            \toprule
            \textbf{Scheduler} & \textbf{FID50k$\downarrow$} & \textbf{IS$\uparrow$} \\ \midrule
            root   & 39.08 & 32.55 \\
            linear & 31.29 & 38.02 \\
            cosine & 31.45 & 36.11 \\
            square & 31.27 & 37.79 \\
            arccos & \textbf{29.68} & \textbf{40.28} \\
            \bottomrule
        \end{tabular}
        \caption{Ablation on \textcolor{darkorange}{scheduling methods} (\(\alpha=0.2\), 16 steps). Convex schedulers like arccos perform best.}
        \label{tab:scheduling_methods}
    \end{minipage}
    \hfill
    \begin{minipage}{0.48\textwidth}  %
        \centering
        \begin{tabular}{l|cc}
            \toprule
            \textbf{Sequence} & \textbf{FID50k$\downarrow$} & \textbf{IS$\uparrow$} \\ \midrule
            Halton        & 31.62 & \textbf{37.87} \\
            Halton + Roll & \textbf{31.27} & 37.79 \\
            Raster Scan   & 43.60 & 34.29 \\
            Spiral        & 36.34 & 26.71 \\
            \bottomrule
        \end{tabular}
        \caption{Ablation on \textcolor{olive}{token ordering} on ImageNet 256. Uniform prediction (Halton sequence) improves generation.}
        \label{tab:ablation_init}
    \end{minipage}
\end{table}

\paragraph{Token Prediction Order}
We compare different token selection strategies $\{\text{Halton, Spiral, Raster Scan}\}$ in Table~\ref{tab:ablation_init}. We find that Halton ordering significantly outperforms raster scan and spiral selection in both metrics. This demonstrates the advantage of structured, but detached, token sampling in guiding the prediction process more effectively. We also compare the performance of ‘starting from the same token’ location versus ‘rolling out the sequence’ (Halton+Roll). Specifically, we apply a circular shift to the sequence during both training and testing, enabling the model to begin from any token location in the image. Our results indicate that there is no significant boost in performance between those two strategies.

\begin{figure*}[ht]
    \centering
    \begin{subfigure}[b]{0.45\textwidth}
        \centering
        \includegraphics[width=\textwidth]{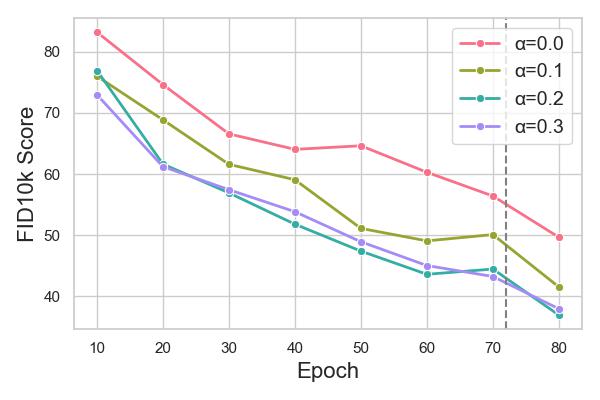}
        \caption{Ablation on the \textcolor{teal}{random token injection $\alpha$}}
        \label{fig:ablation_alpha}
    \end{subfigure}
    \hfill
    \begin{subfigure}[b]{0.45\textwidth}
        \centering
        \includegraphics[width=\textwidth]{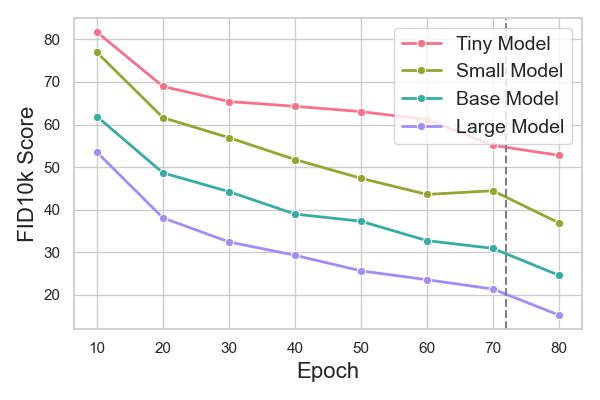}
        \caption{Ablation on \textcolor{violet}{model sizes}}
        \label{fig:ablation_model}
    \end{subfigure}
    \\
    \begin{subfigure}[b]{0.45\textwidth}
        \centering
        \includegraphics[width=\textwidth]{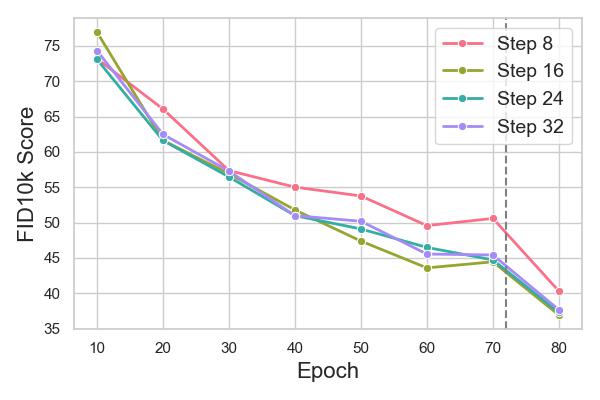}
        \caption{Ablation on \textcolor{blue}{prediction steps}}
        \label{fig:ablation_step}
    \end{subfigure}
    \hfill
    \begin{subfigure}[b]{0.45\textwidth}
        \centering
        \includegraphics[width=\textwidth]{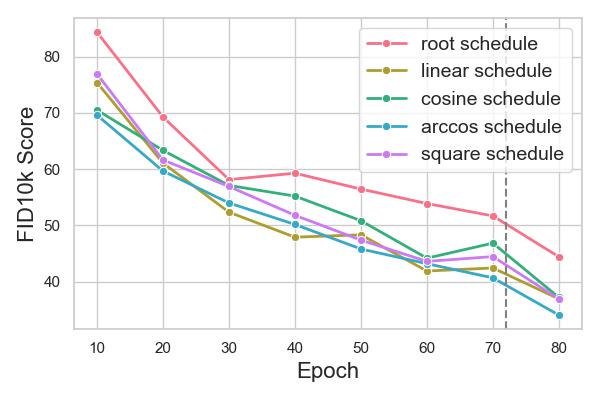}
        \caption{Ablation on \textcolor{darkorange}{scheduling methods}}
        \label{fig:scheduling_methods}
    \end{subfigure}
    \caption{\textbf{ImageNet 256:} FID10k evolution across model training on ImageNet-256, without cfg and 410k iterations. The moment where learning rate decay was applied is showcased by the dash grey line.}
    \label{fig:ablation}
    
\end{figure*}

\paragraph{Summary of Findings:}
Our ablation study highlights key insights into the impact of different hyper-parameters on model performance~\autoref{fig:ablation}. We find that using a moderate level of \textcolor{teal}{random token injection} ($\alpha=0.2$) drastically improves the performance. Setting the number of \textcolor{blue}{prediction steps} to between $16$ and $32$ provides an optimal trade-off between efficiency and quality. Additionally, increasing the number of tokens following an arccos-based \textcolor{darkorange}{scheduling strategy} outperforms alternative approaches for guiding token prediction. Finally, leveraging the Halton sequence for the \textcolor{olive}{token ordering} leads to significantly enhanced image quality and therefore it serves as the baseline method we compare to in the main paper.

\section{Sampling pattern}\label{sec:sampling_patern}

\begin{figure}[ht]
    \centering
    \begin{subfigure}[b]{0.48\columnwidth}
        \centering 
        \includegraphics[width=0.7\linewidth]{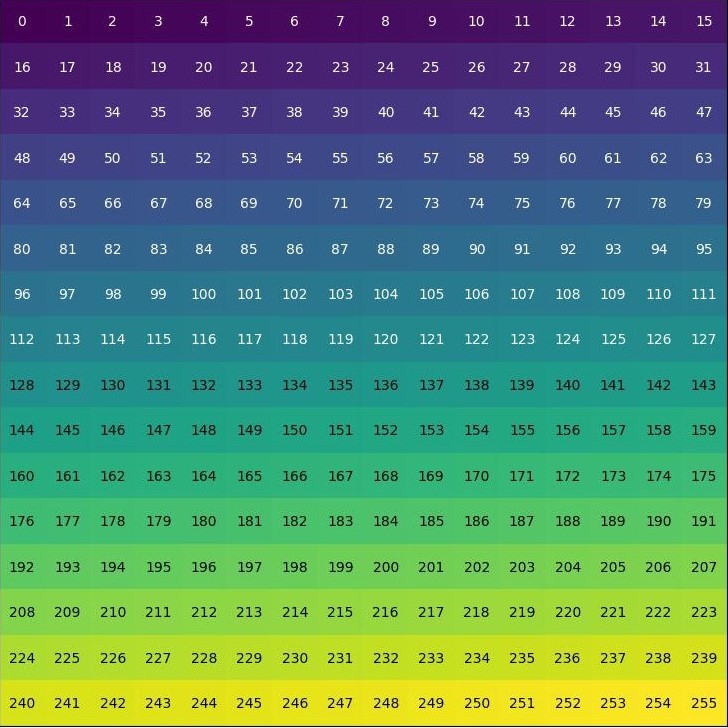}
        \caption{Raster Scan in 2D}
    \end{subfigure}
    \hspace{0.02\columnwidth}
    \begin{subfigure}[b]{0.48\columnwidth}
        \centering
        \includegraphics[width=0.7\linewidth]{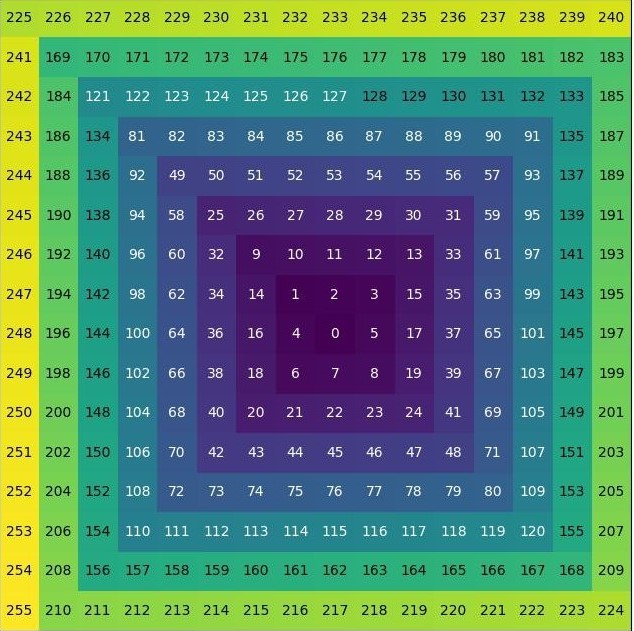}
        \caption{Spiral Sequence in 2D}
    \end{subfigure}

    \vspace{0.5em}
    \begin{subfigure}[b]{0.48\columnwidth}
        \centering
        \includegraphics[width=0.7\linewidth]{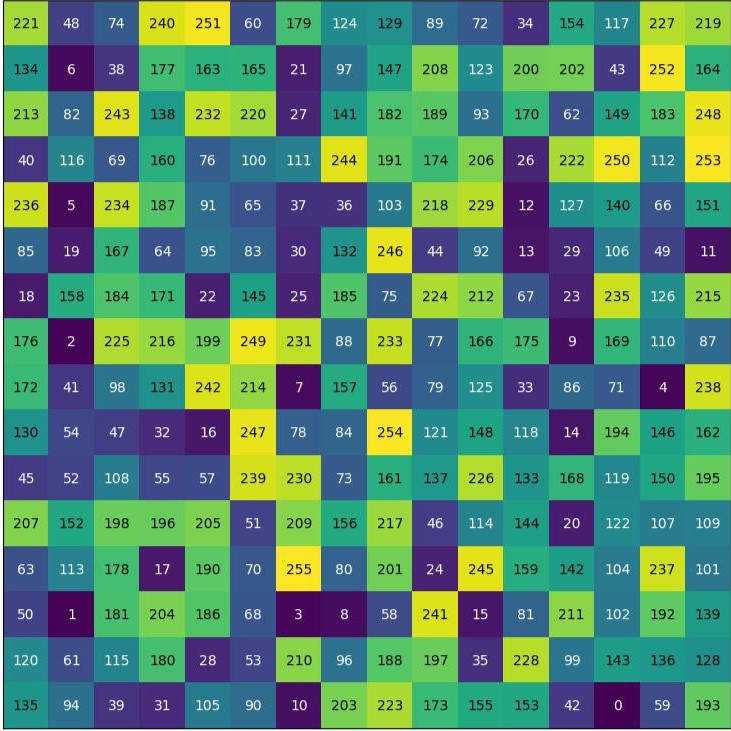}
        \caption{Random Sequence in 2D}
    \end{subfigure}
    \hspace{0.02\columnwidth}
    \begin{subfigure}[b]{0.48\columnwidth}
        \centering
        \includegraphics[width=0.7\linewidth]{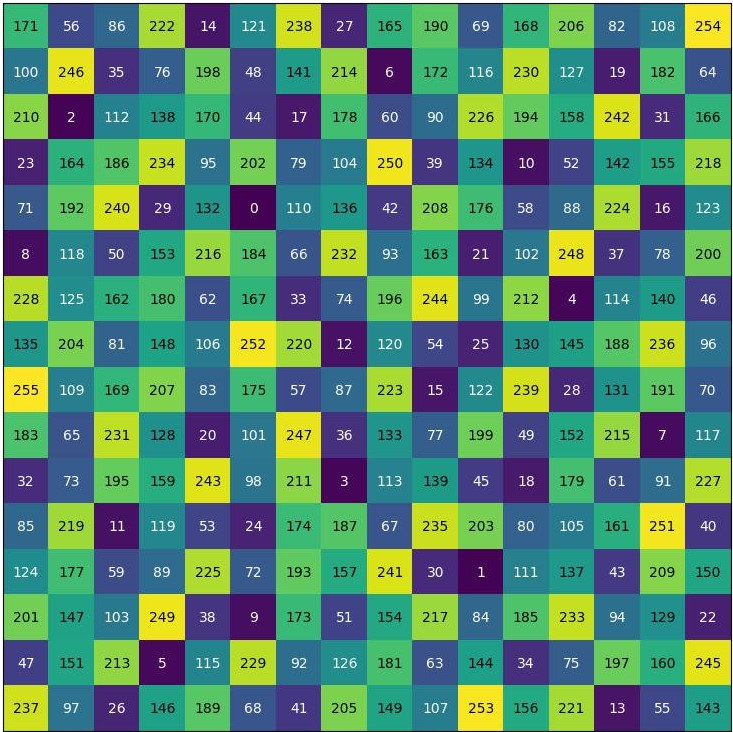}
        \caption{Halton Sequence in 2D}
    \end{subfigure}

    \caption{Visualization of different sequence orderings in 2D.}
    \label{fig:sequence_order}
\end{figure}

An important aspect of our method is the order in which tokens are predicted. In the previous section, we show that the Halton ordering outperforms alternative approaches. \autoref{fig:sequence_order} illustrates these token sequences on a \(16 \times 16\) grid. Notably, the Raster scan method immediately reveals its limitations when applied to a 2D grid, as it enforces a rigid left-to-right, top-to-bottom structure. In contrast, both the Random and Halton sequences achieve a more uniform distribution across the grid, avoiding biases toward specific regions. Compared to random ordering, the Halton sequence is more robust to gaps; for instance, in the random ordering example below, the 17th token is sampled early, while its neighboring tokens are selected much later, leading to a less balanced and structured prediction process.

\section{Hyper-parameters}\label{sec:hparam}
\begin{figure}[ht]
    \centering
    \begin{subfigure}[b]{0.48\columnwidth}
        \centering
        \includegraphics[width=\linewidth]{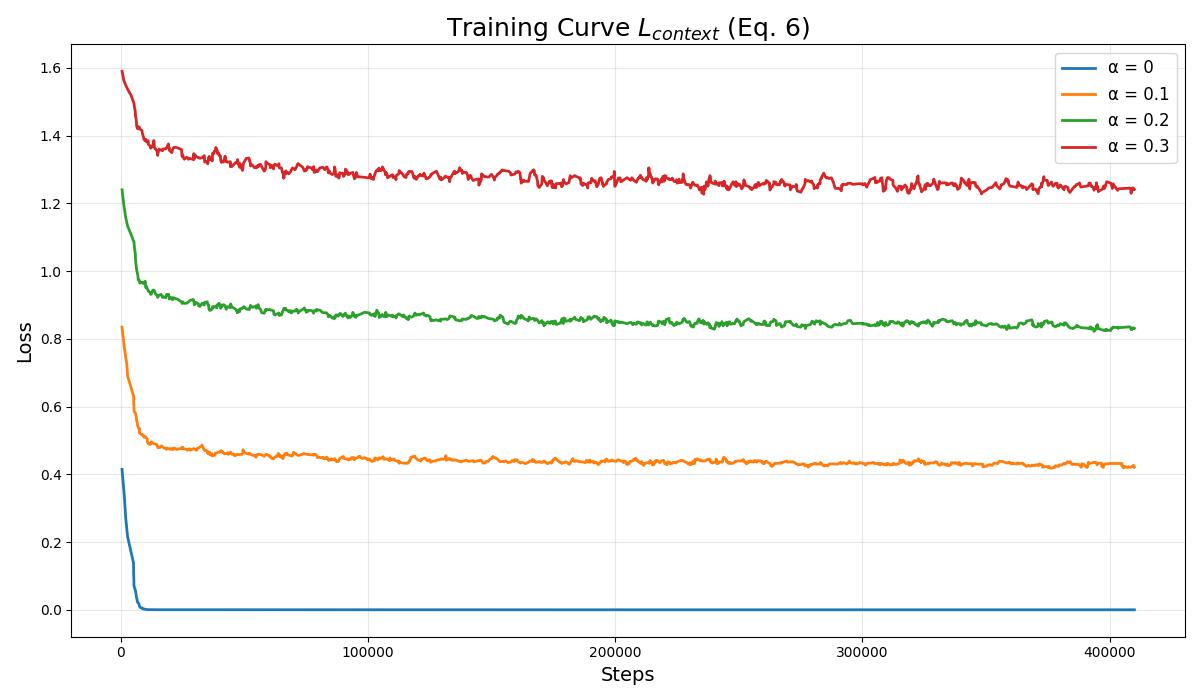}
        \caption{Loss of Eq. 6}
    \end{subfigure}
    \begin{subfigure}[b]{0.48\columnwidth}
        \centering 
        \includegraphics[width=\linewidth]{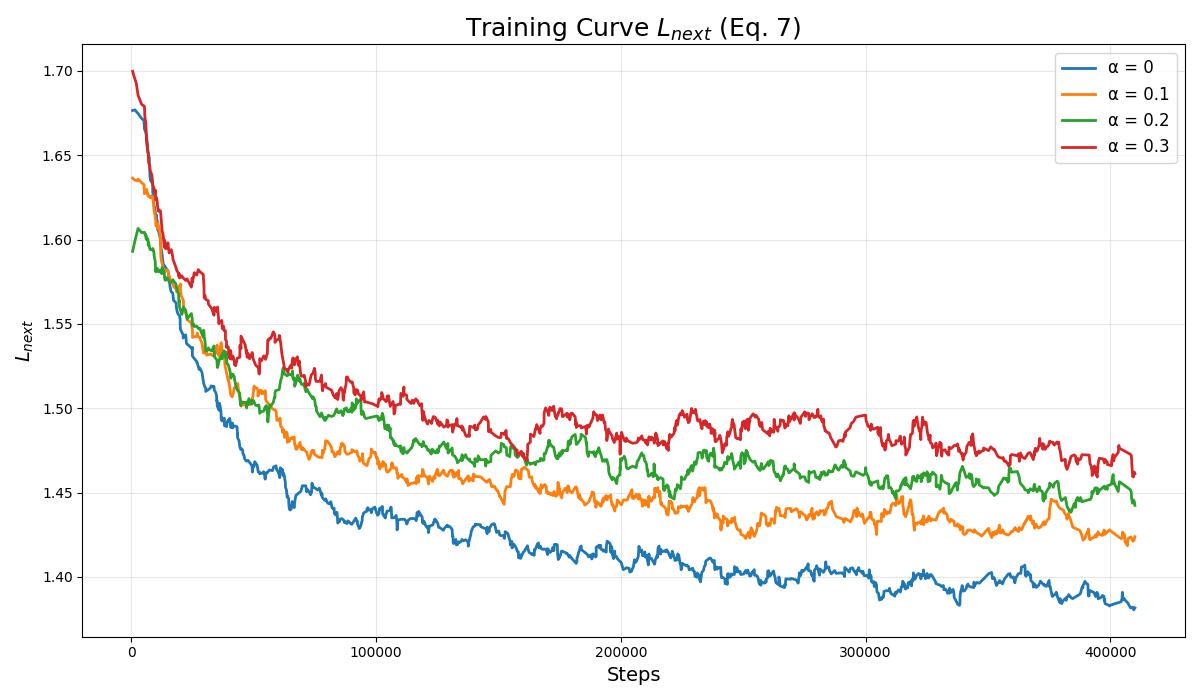}
        \caption{Loss of Eq. 7}
    \end{subfigure}
    \hspace{0.02\columnwidth}

    \caption{Training Loss of \ours{}-Small across different values of $\alpha$}
    \label{fig:training_losses}
\end{figure}

In Table~\ref{tab:hparams_table}, we provide the hyper-parameters used for training our class-to-image and class-to-video models on CIFAR-10, ImageNet, UCF101 and NuScenes datasets. The training process differs primarily in the number of steps and batch sizes, reflecting the scale of each dataset. A cosine learning rate decay is applied only for the last 10\% of the iterations and we use 2,500 warmup steps to stabilize early training. We incorporate gradient clipping (norm = 1) to prevent exploding gradients and classifier-free guidance (CFG) dropout of 0.1 for better sample diversity. The CIFAR-10 model applies horizontal flip augmentation, while no data augmentation is used for ImageNet. Both models are trained using bfloat16 precision for computational efficiency. These hyper-parameters were chosen to ensure stable training while balancing efficiency and performance across different datasets. Finally, we sweep our model size according to~\autoref{tab:transformer_config}. We also show the training dynamics of the two losses in~\autoref{fig:training_losses}

\begin{table}[ht]
    \centering
    \begin{tabular}{lccccc}
        \toprule
        \textbf{Model}   & \textbf{Parameters} & \textbf{GFLOPs} & \textbf{Heads} & \textbf{Hidden Dim} & \textbf{Width} \\\midrule
        \ours{}-Tiny     & 24M   & 4.0   & 6  & 384  & 6  \\
        \ours{}-Small    & 50M   & 9.0   & 8  & 512  & 8  \\
        \ours{}-Base     & 143M  & 25.0  & 12 & 768  & 12 \\
        \ours{}-Large    & 480M  & 83.0  & 16 & 1024 & 24 \\
        \ours{}-XLarge   & 693M  & 119.0 & 16 & 1152 & 28 \\
        \bottomrule
    \end{tabular}
    
    \caption{Transformer model configurations for \(16 \times 16\) input size.}
    \label{tab:transformer_config}
\end{table}

\begin{table}[ht]
    \centering
    \resizebox{\linewidth}{!}{
    \begin{tabular}{ccccc} \toprule
       \textbf{Condition}  & \multicolumn{1}{c}{\textbf{Cifar10}} & \multicolumn{1}{c}{\textbf{ImageNet}} & \multicolumn{1}{c}{\textbf{UCF101}} & \multicolumn{1}{c}{\textbf{NuScenes}} \\
       \midrule
       Training steps     & $400k$                        & $1.5M$                        & $410k$                       & $410k$                       \\ 
       Batch size         & $128$                         & $256$                         & $256$                        & $8$                          \\
       Learning rate      & $1 \times 10^{-4}$            & $1 \times 10^{-4}$            & $1 \times 10^{-4}$           & $2 \times 10^{-4}$           \\
       Weight decay       & $0.03$                        & $0.03$                        & $0.03$                       & $0.03$                       \\
       Optimizer          & AdamW                         & AdamW                         & AdamW                        & AdamW                        \\
       Momentum           & $\beta_1=0.9, \beta_2=0.999$  & $\beta_1=0.9, \beta_2=0.999$  & $\beta_1=0.9, \beta_2=0.999$ & $\beta_1=0.9, \beta_2=0.999$  \\
       Lr scheduler       & Cosine                        & Cosine                        & Cosine                       & Cosine                        \\
       Warmup steps       & $2500$                        & $2500$                        & $2500$                       & $2500$                        \\
       Gradient clip norm & $1$                           & $1$                           & $1$                          & $1$                           \\
       CFG dropout        & $0.1$                         & $0.1$                         & $0.1$                        & $0.1$                         \\ 
       dropout            & $0.1$                         & $0.1$                         & $0.1$                        & $0.1$                         \\ 
       Data aug.          & Horizontal Flip               & No                            & No                           & No                            \\
       Precision          & bf16                          & bf16                          & bf16                         & bf16                          \\
       \bottomrule
    \end{tabular}}
    \caption{Hyper-parameters used in the training of class-to-image and class-to-video models.}
    \label{tab:hparams_table}
\end{table}

\section{Inference Speed Analysis}\label{sec:speed}
We measured inference speed for each model across multiple decoding steps (the mean over 1000 samples with batch size of 8*2 for CFG). From these runs, we report the steps and the throughput (images/sec). The experiment is made on a Nvidia A100, bfloat16 precision and cfg. We do not use KV-cache or any other tricks. The correction does not cause any overhead, but could prevent KV-caching. We report those values in~\autoref{tab:images_per_sec_merged}.

\begin{table}[ht]
\centering
\setlength{\tabcolsep}{8pt}
\begin{tabular}{lcccc}
\toprule
\textbf{Model} & \textbf{Step} & \textbf{input size 16$\times$16} & \textbf{input size 24$\times$24} \\
& & (img/s) &(img/s) \\
\midrule
\multirow{4}{*}{\textbf{\ours{}-Tiny}} 
 & 8  & 80.63 & 41.98 \\
 & 16 & 55.89 & 30.32 \\
 & 24 & 42.53 & 23.48 \\
 & 32 & 34.45 & 19.30 \\
\midrule
\multirow{4}{*}{\textbf{\ours{}-Small}}
 & 8  & 71.66 & 37.28 \\
 & 16 & 47.18 & 25.51 \\
 & 24 & 35.30 & 19.39 \\
 & 32 & 28.21 & 15.61 \\
\midrule
\multirow{4}{*}{\textbf{\ours{}-Base}} 
 & 8  & 54.03 & 27.17 \\
 & 16 & 33.39 & 16.87 \\
 & 24 & 24.16 & 12.21 \\
 & 32 & 18.88 & 9.58 \\
\midrule
\multirow{4}{*}{\textbf{\ours{}-Large}}
 & 8  & 28.18 & 14.37 \\
 & 16 & 15.53 & 7.98 \\
 & 24 & 10.73 & 5.53 \\
 & 32 & 8.21 & 4.22 \\
\midrule
\multirow{4}{*}{\textbf{\ours{}-XLarge}}
 & 8  & 21.93 & 10.74 \\
 & 16 & 11.90 & 5.80 \\
 & 24 & 8.15 & 3.97 \\
 & 32 & 6.18 & 3.01 \\
\bottomrule
\end{tabular}
\caption{Throughput comparison (img/s) for different input sizes.}
\label{tab:images_per_sec_merged}
\end{table}

\section{Limitations} \label{sec:limitation}
Like other auto-regressive approaches, \ours{} requires a fixed token unmasking schedule defined at training and maintained at inference, which limits flexibility during prediction. While not an inherent limitation of the method, our experiments were restricted to models below 1 billion parameters to keep computational costs manageable. Extending \ours{} to large-scale image and video generation remains future work.

\section{Prompts}
\begin{figure}[t]
    \centering

    \includegraphics[width=0.155\linewidth]{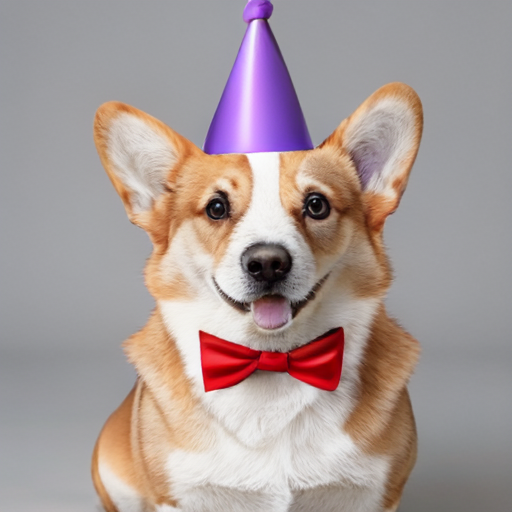}\hspace{-0.15cm}
    \includegraphics[width=0.155\linewidth]{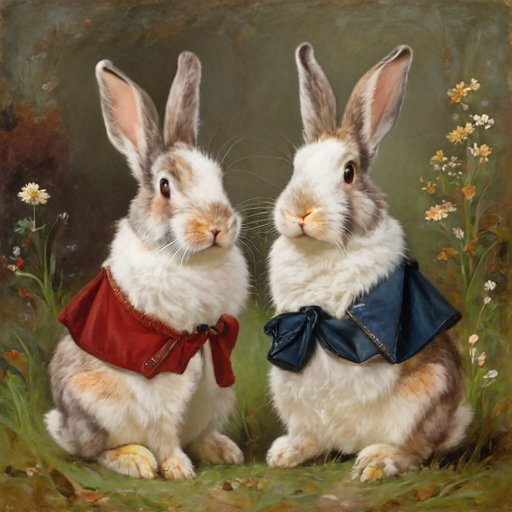}\hspace{-0.15cm}
    \includegraphics[width=0.155\linewidth]{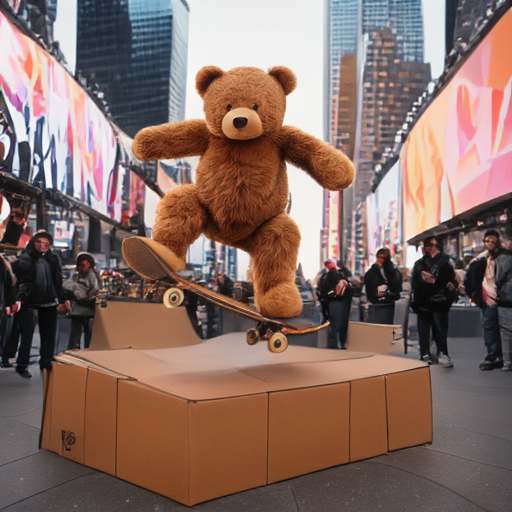}\hspace{-0.15cm}
    \includegraphics[width=0.155\linewidth]{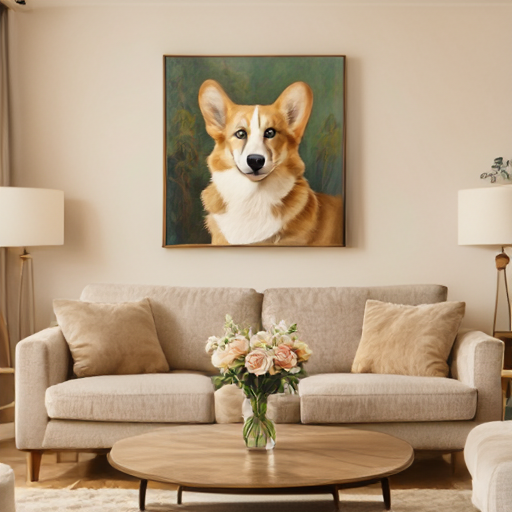}\hspace{-0.15cm}
    \includegraphics[width=0.155\linewidth]{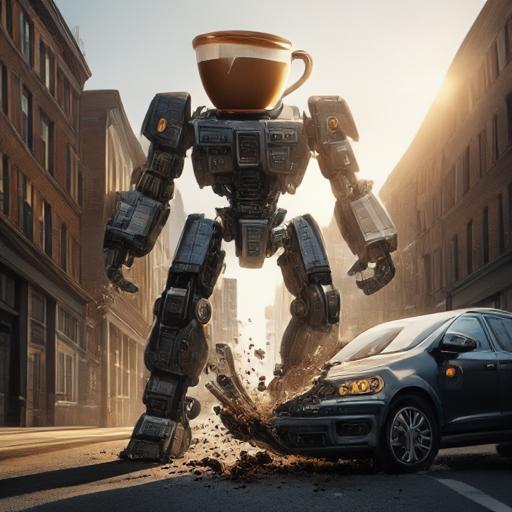}\hspace{-0.15cm}
    \includegraphics[width=0.155\linewidth]{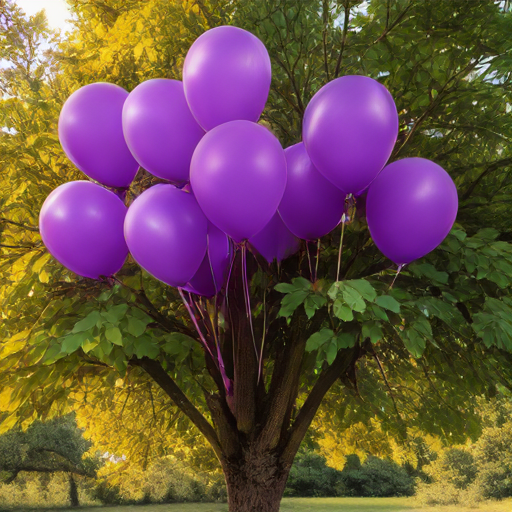}\\
    \includegraphics[width=0.155\linewidth]{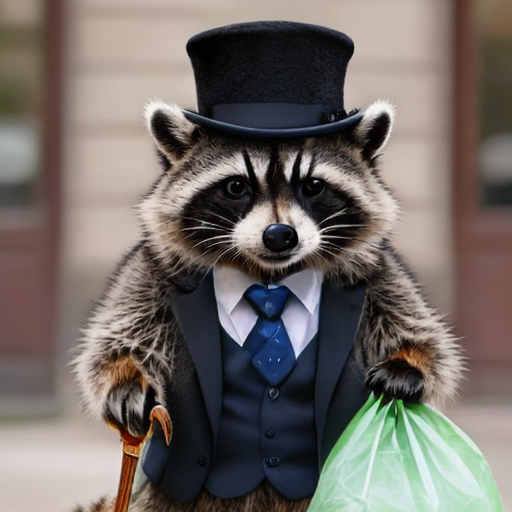}\hspace{-0.15cm}
    \includegraphics[width=0.155\linewidth]{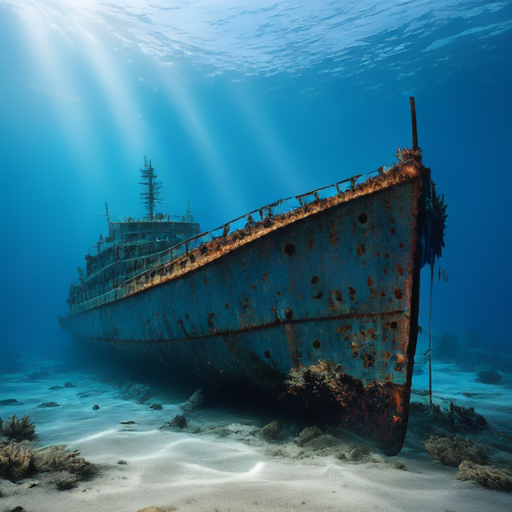}\hspace{-0.15cm}
    \includegraphics[width=0.155\linewidth]{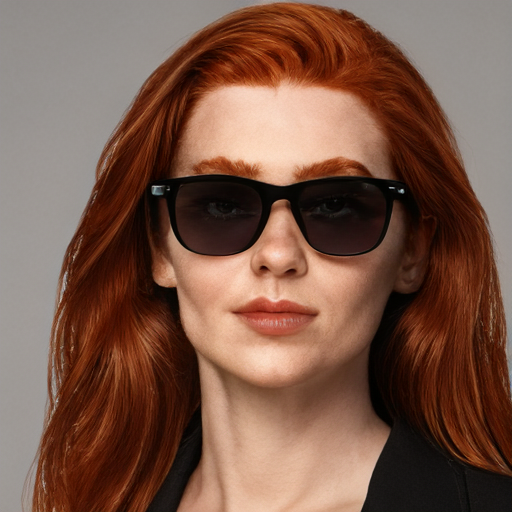}\hspace{-0.15cm}
    \includegraphics[width=0.155\linewidth]{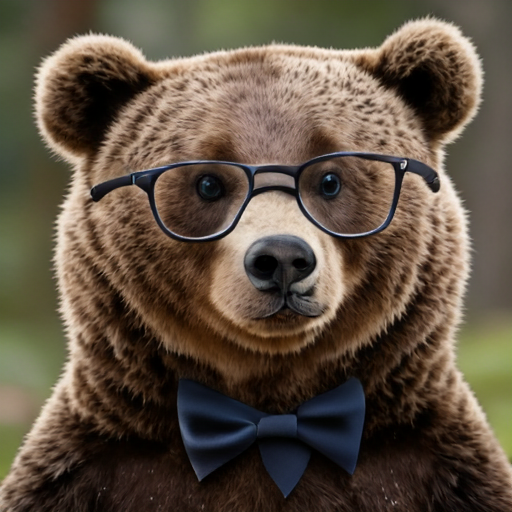}\hspace{-0.15cm}
    \includegraphics[width=0.155\linewidth]{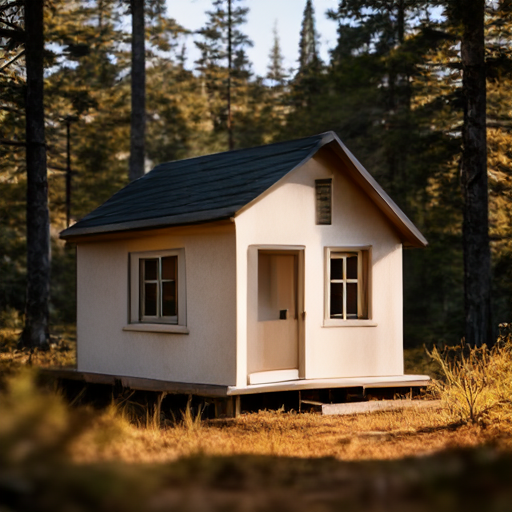}\hspace{-0.15cm}
    \includegraphics[width=0.155\linewidth]{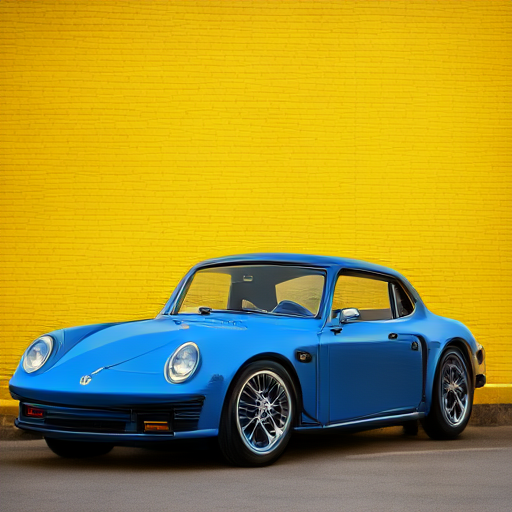}\\
    \includegraphics[width=0.155\linewidth]{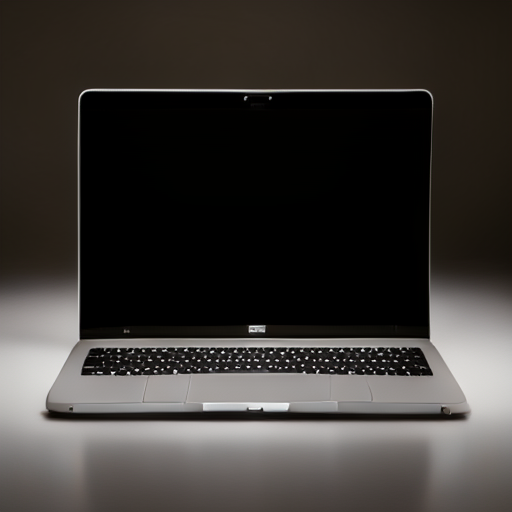}\hspace{-0.15cm}
    \includegraphics[width=0.155\linewidth]{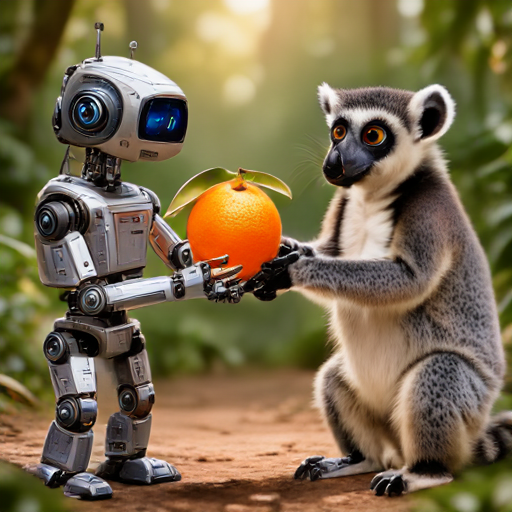}\hspace{-0.15cm}
    \includegraphics[width=0.155\linewidth]{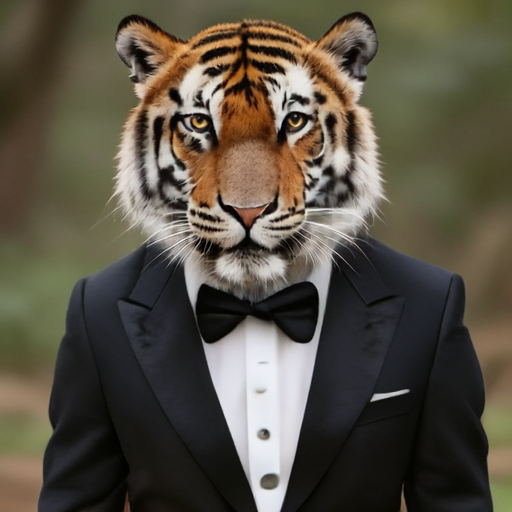}\hspace{-0.15cm}
    \includegraphics[width=0.155\linewidth]{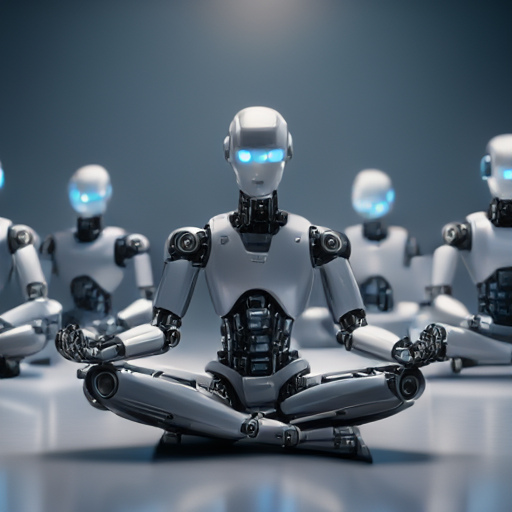}\hspace{-0.15cm}
    \includegraphics[width=0.155\linewidth]{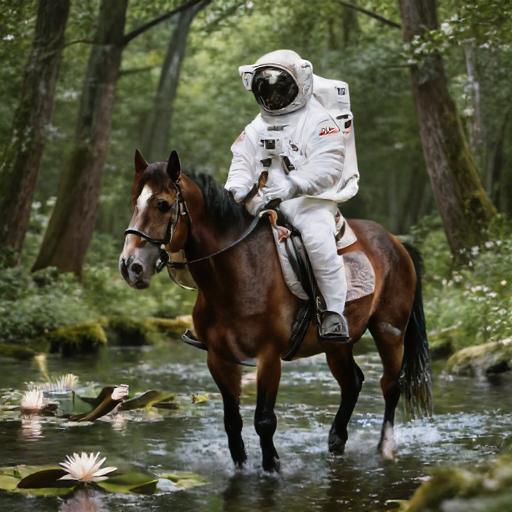}\hspace{-0.15cm}
    \includegraphics[width=0.155\linewidth]{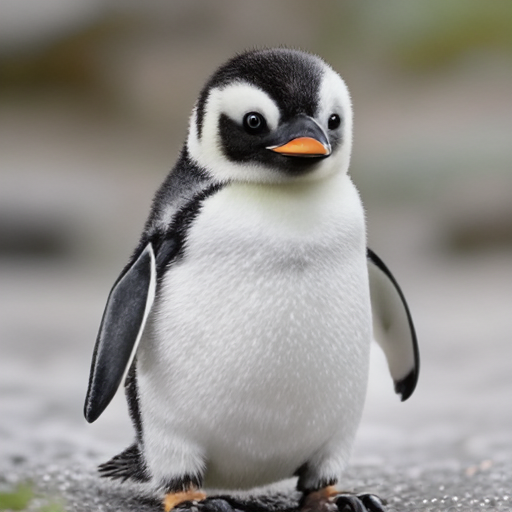}\\
    \includegraphics[width=0.155\linewidth]{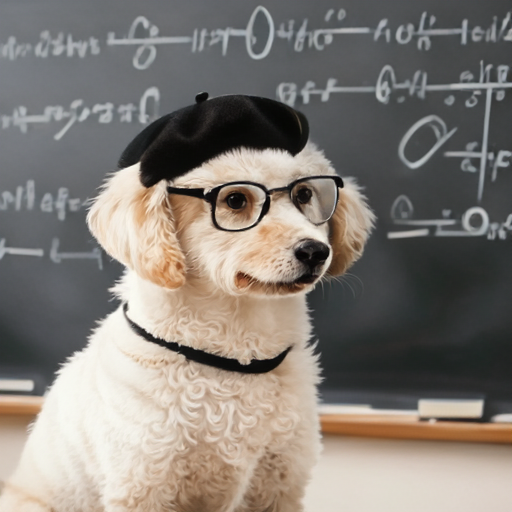}\hspace{-0.15cm}
    \includegraphics[width=0.155\linewidth]{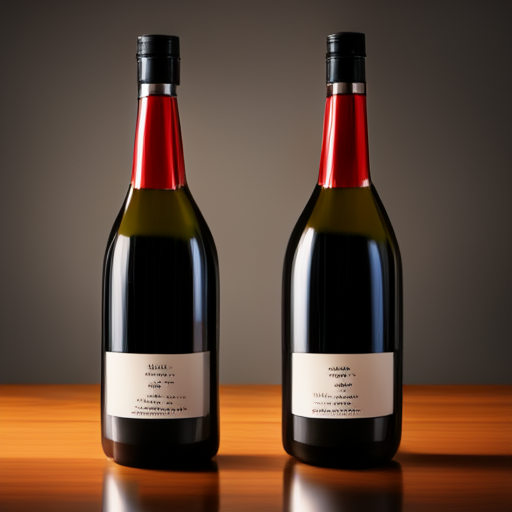}\hspace{-0.15cm}
    \includegraphics[width=0.155\linewidth]{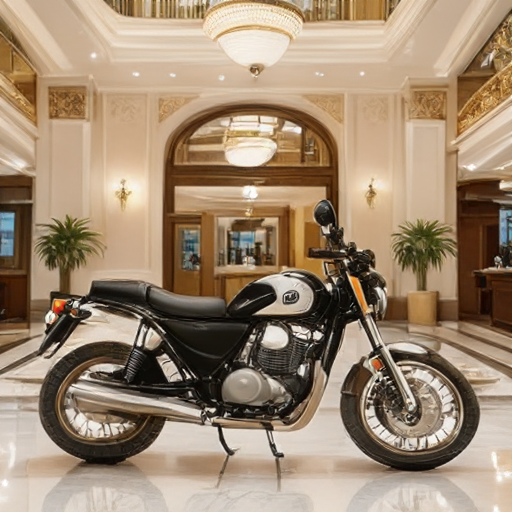}\hspace{-0.15cm}
    \includegraphics[width=0.155\linewidth]{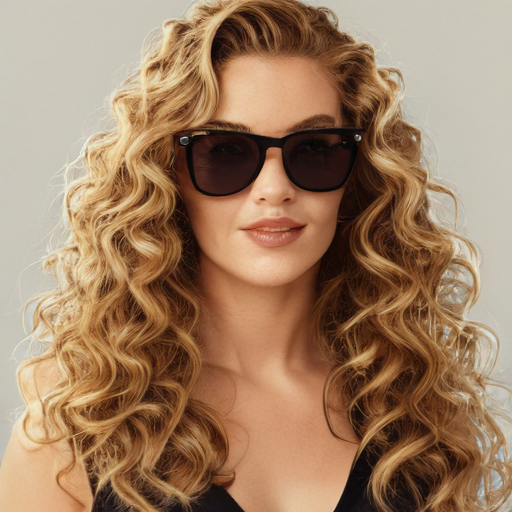}\hspace{-0.15cm}
    \includegraphics[width=0.155\linewidth]{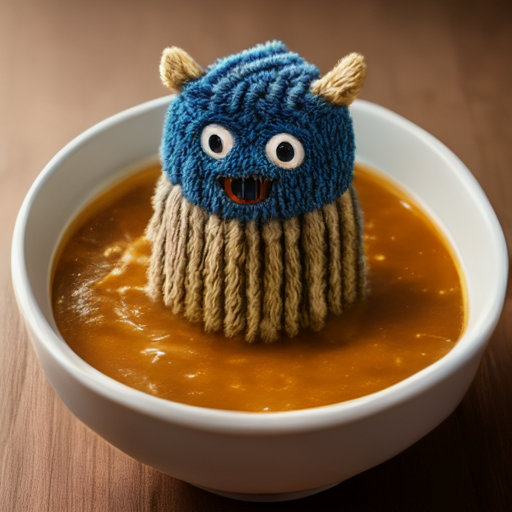}\hspace{-0.15cm}
    \includegraphics[width=0.155\linewidth]{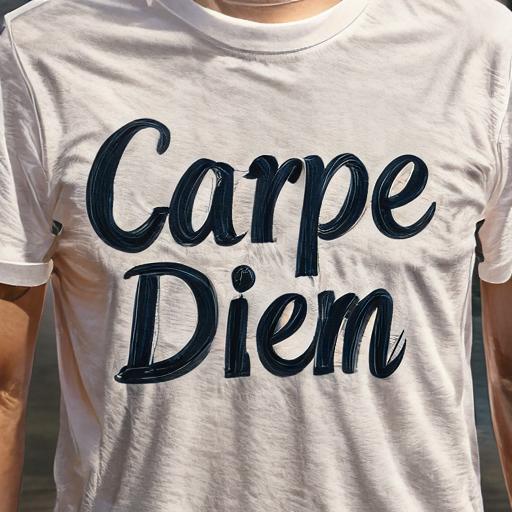}

    \caption{\textbf{Randomly generated samples from PartiPrompt captions.}
    \ours{}-XLarge generates diverse high-quality images across a broad range of concepts and visual styles, demonstrating strong text-image alignment, realistic details, and compositional diversity.}
    \label{fig:qualitative_txt2img_parti}
\end{figure}

For~\autoref{fig:teaser_t2i}, the caption used to generate the images are:
{\small
\begin{itemize}
    \item \textbf{Image 1:} "In a library bathed in ethereal light, two figures adorned with translucent jellyfish-like domes hover above their heads, their tendrils intertwining with delicate, glowing threads as they tend to a book that seems to pulse with life, defying the boundaries between the mundane and the magical."
    \item \textbf{Image 2:} "A lone astronaut drifts serenely beneath the surface of an ethereal ocean, suspended amidst a cascade of glowing, translucent threads that seem to weave the fabric of a dreamworld, while scattered fragments of light dance around like remnants of forgotten memories."
    \item \textbf{Image 3:} "A close-up portrait of a person's face, but it's heavily distorted with severe pixelation and vibrant color banding, sections of the face are misaligned and shifted, creating a fractured and abstract look, glitch art."
    \item \textbf{Image 4:} "A tranquil lakeside scene at dawn, with misty mountains in the background and a lone wooden boat floating on the still water, all rendered in a soft, ethereal watercolor illustration style with visible paper texture."
    \item \textbf{Image 5:} "A vibrant street wall covered in colorful graffiti, the centerpiece spells 'HALTON', in a storm of colors."
    \item \textbf{Image 6:} "A photorealistic macro shot of two open books resting on a mossy forest floor with scattered autumn leaves. Detailed origami birds appear to take flight directly from the open pages—a white bird in the sharp foreground and a pale purple bird in the slightly blurred background. The setting is a mystical, softly out-of-focus forest with tall trees and warm, dappled sunlight, evoking a whimsical and magical atmosphere."
    \item \textbf{Image 7:} "A robot cooking in the kitchen."
    \item \textbf{Image 8:} "A photo of a teddy bear made of water."
    \item \textbf{Image 9:} "Two cups of coffee, one with latte art of the words 'LOVE' written in one. The other has latte art of the words 'PEACE' written in the other."
    \item \textbf{Image 10:} "A pixel art corgi pizza."
    \item \textbf{Image 11:} "A cosmic teardrop carved from the desert sand cradles a swirling galaxy, defying gravity as it floats serenely against a starlit sky."
    \item \textbf{Image 12:} "a mixed media image with a photograph of a woman with long orange hair over a background that is a sketch of a city skyline."
\end{itemize}
}
For~\autoref{fig:qualitative_txt2img_parti}, the caption used to generate the images are:
\begin{itemize}
    \item \textbf{Image 1:}  "A corgi wearing a red bowtie and a purple party hat."
    \item \textbf{Image 2:}  "An oil painting of two rabbits in the style of American Gothic, wearing the same clothes as in the original."
    \item \textbf{Image 3:}  "A teddybear on a skateboard in Times Square, doing tricks on a cardboard box ramp."
    \item \textbf{Image 4:}  "A cozy living room with a painting of a corgi on the wall above a couch and a round coffee table in front of a couch and a vase of flowers on a coffee table."
    \item \textbf{Image 5:}  "A massive robot with a coffee cup head. It is standing in the street with one foot smashing a car."
    \item \textbf{Image 6:}  "A tree with leaves that look like purple balloons."
    \item \textbf{Image 7:}  "A raccoon wearing formal clothes, wearing a tophat and holding a cane. The raccoon is holding a garbage bag."
    \item \textbf{Image 8:}  "A sunken ship at the bottom of the ocean."
    \item \textbf{Image 9:} "A woman with sunglasses and red hair."
    \item \textbf{Image 10:}  "A nerdy bear wearing glasses and a bowtie."
    \item \textbf{Image 11:}  "A small house in the wilderness."
    \item \textbf{Image 12:}  "A blue Porsche 356 parked in front of a yellow brick wall."
    \item \textbf{Image 13:}  "A laptop."
    \item \textbf{Image 14:}  "A robot gives a wombat an orange and a lemur a banana."
    \item \textbf{Image 15:}  "A tiger wearing a tuxedo."
    \item \textbf{Image 16:}  "Robots meditating."
    \item \textbf{Image 17:}  "A photo of an astronaut riding a horse in the forest. There is a river in front of them with water lilies."
    \item \textbf{Image 18:}  "A baby penguin."
    \item \textbf{Image 19:}  "A cream-colored labradoodle wearing glasses and black beret teaching calculus at a blackboard."
    \item \textbf{Image 20:}  "Two bottles of wines."
    \item \textbf{Image 21:}  "A motorcycle parked in an ornate bank lobby."
    \item \textbf{Image 22:}  "A girl with long curly blonde hair and sunglasses."
    \item \textbf{Image 23:}  "A bowl of soup that looks like a monster knitted out of wool."
    \item \textbf{Image 24:}  "a t-shirt with Carpe Diem written on it."
\end{itemize}

\section{Additional qualitative results}\label{sec:qual_res}
\begin{figure*}[ht]
    \centering
    \begin{subfigure}{0.45\textwidth}
        \includegraphics[width=\linewidth]{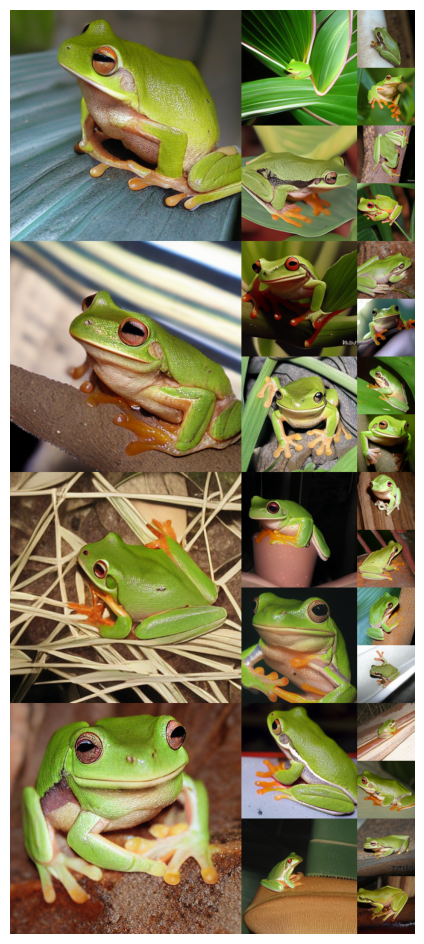}
        \caption{Tree frog \imagenetid{031}}
        \label{fig:tree_frog}
    \end{subfigure}
    \hfill
    \begin{subfigure}{0.45\textwidth}
        \includegraphics[width=\linewidth]{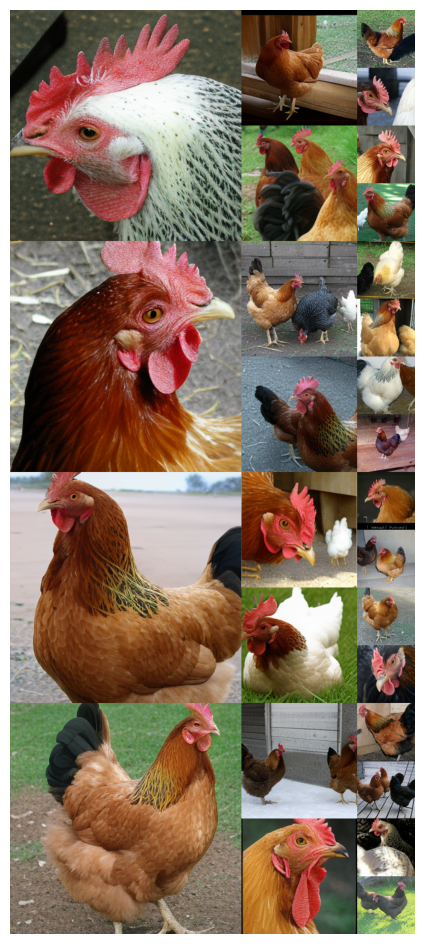}
        \caption{Chicken \imagenetid{008}}
        \label{fig:chicken}
    \end{subfigure}
    \caption{Random samples from our \ours{}-Large model.\\ $\alpha=0.2$, $cfg=4.0$ and $24$ steps.}
    \label{fig:viz1}
\end{figure*}

\begin{figure*}[ht]
    \centering
    \begin{subfigure}{0.45\textwidth}
        \includegraphics[width=\linewidth]{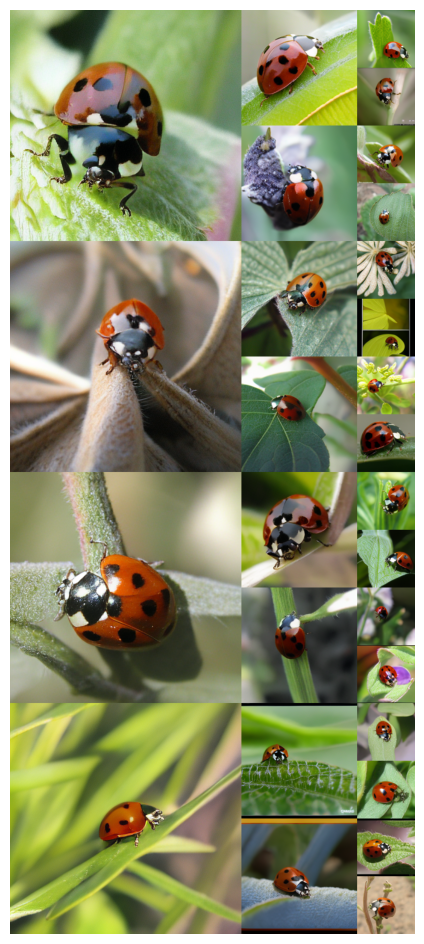}
        \caption{LadyBug \imagenetid{301}}
        \label{fig:ladybug}
    \end{subfigure}
    \hfill
    \begin{subfigure}{0.45\textwidth}
        \includegraphics[width=\linewidth]{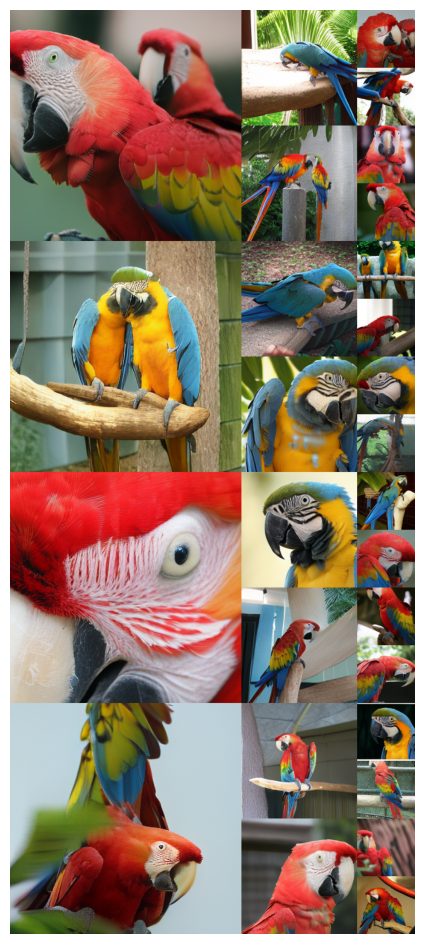}
        \caption{Macaw \imagenetid{88}}
        \label{fig:parrot}
    \end{subfigure}
    \caption{Random samples from our \ours{}-Large model.\\ $\alpha=0.2$, $cfg=4.0$ and $24$ steps.}
    \label{fig:viz2}
\end{figure*}

\begin{figure*}[ht]
    \centering
    \begin{subfigure}{0.45\textwidth}
        \includegraphics[width=\linewidth]{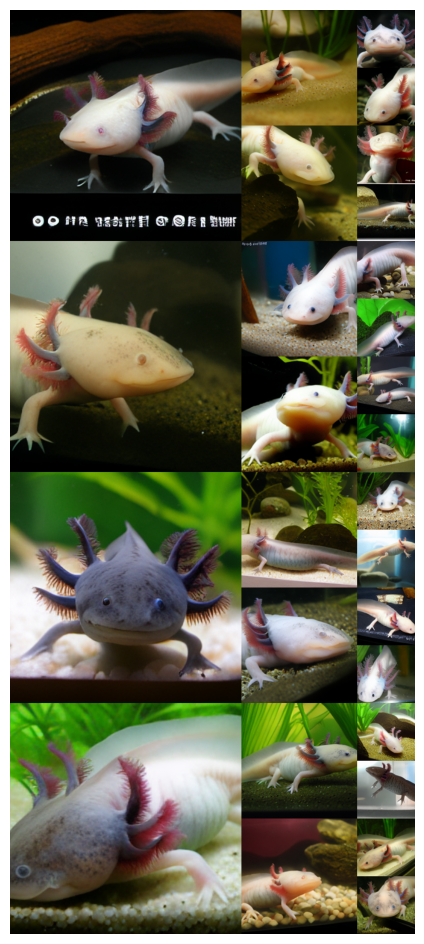}
        \caption{Axolotl \imagenetid{29}}
        \label{fig:axolot}
    \end{subfigure}
    \hfill
    \begin{subfigure}{0.45\textwidth}
        \includegraphics[width=\linewidth]{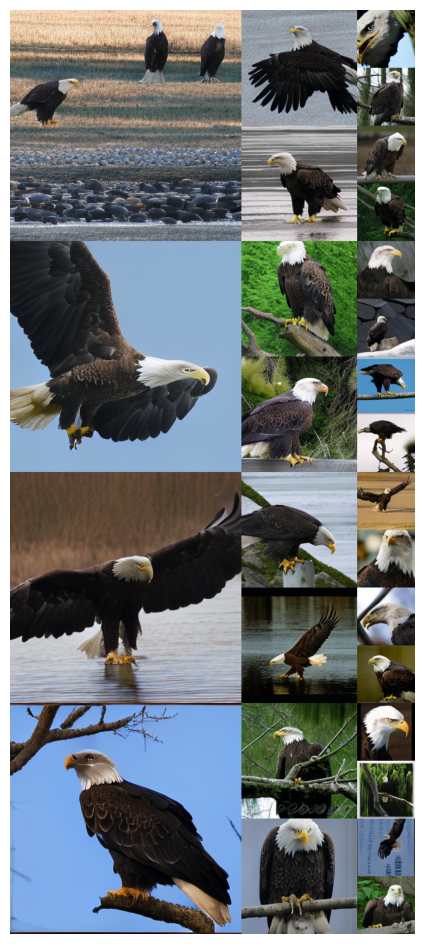}
        \caption{Bald Eagle \imagenetid{22}}
        \label{fig:bald_hawk}
    \end{subfigure}
    \caption{Random samples from our \ours{}-Large model.\\ $\alpha=0.2$, $cfg=4.0$ and $24$ steps.}
    \label{fig:viz3}
\end{figure*}

\begin{figure*}[ht]
    \centering
    \begin{subfigure}{0.45\textwidth}
        \includegraphics[width=\linewidth]{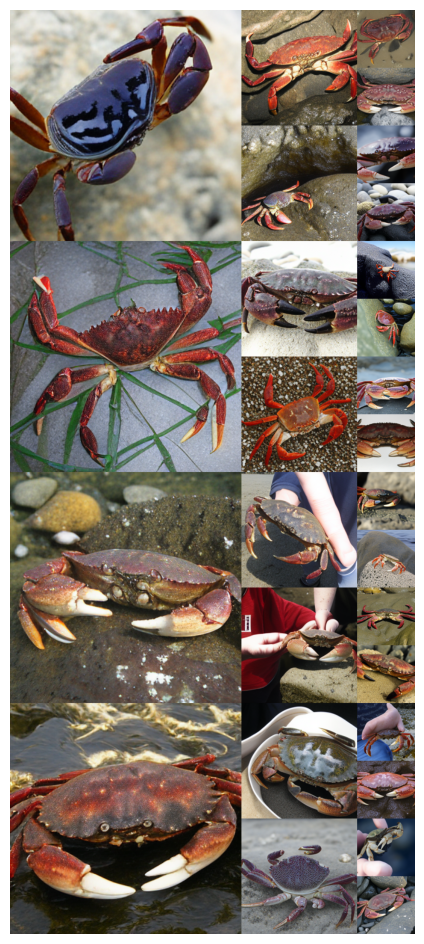}
        \caption{Rock crab \imagenetid{119}}
        \label{fig:rock_crabs}
    \end{subfigure}
    \hfill
    \begin{subfigure}{0.45\textwidth}
        \includegraphics[width=\linewidth]{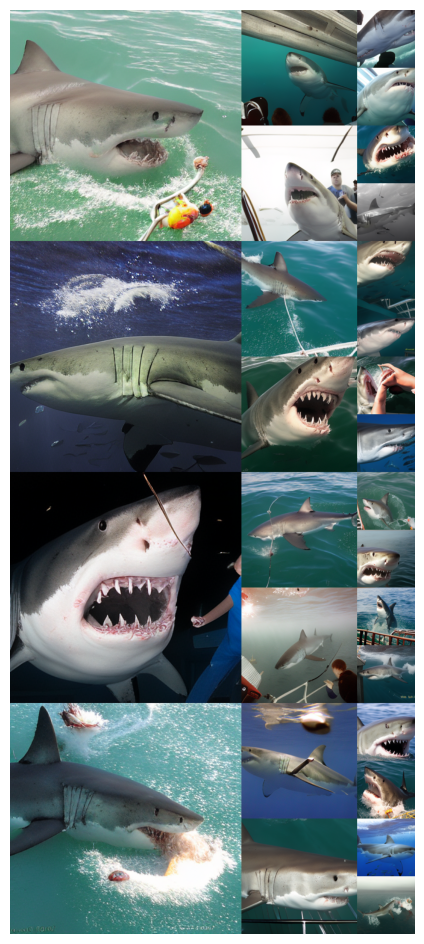}
        \caption{Great white shark \imagenetid{002}}
        \label{fig:great_white_shark}
    \end{subfigure}
    \caption{Random samples from our \ours{}-Large model.\\ $\alpha=0.2$, $cfg=4.0$ and $24$ steps.}
    \label{fig:viz4}
\end{figure*}

\begin{figure*}[ht]
    \centering
    \begin{subfigure}{0.45\textwidth}
        \includegraphics[width=\linewidth]{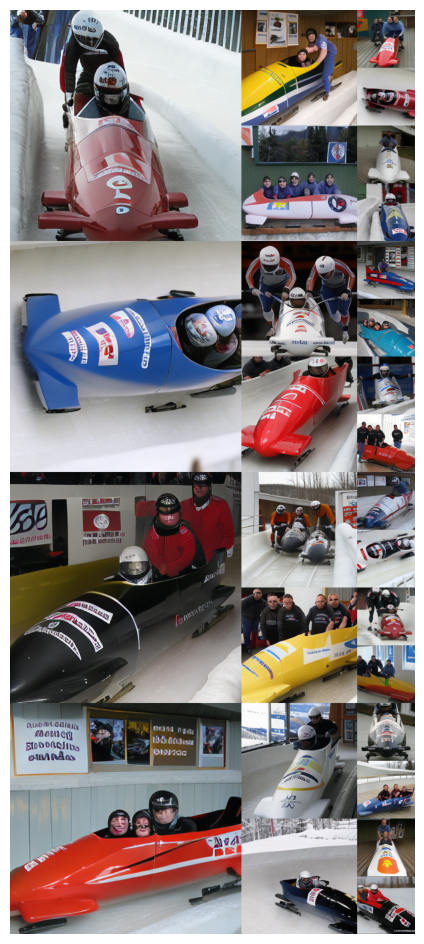}
        \caption{Bobsleigh \imagenetid{450}}
        \label{fig:bobsleigh}
    \end{subfigure}
    \hfill
    \begin{subfigure}{0.45\textwidth}
        \includegraphics[width=\linewidth]{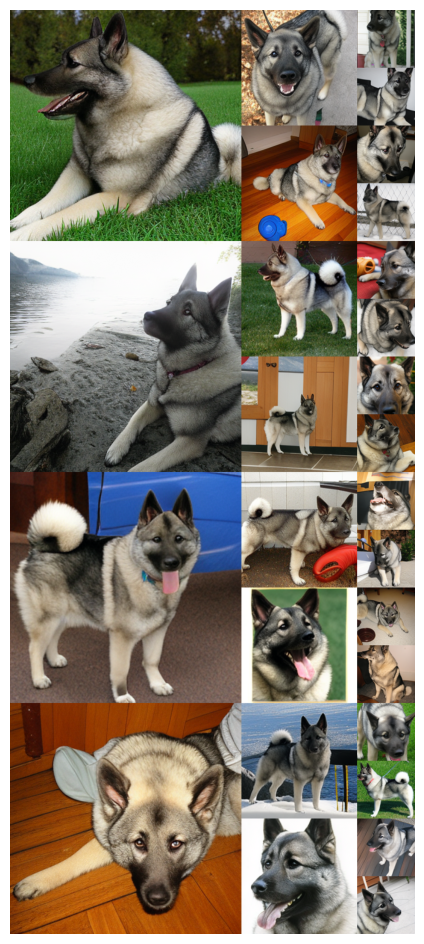}
        \caption{Norwegian Elkhound \imagenetid{174}}
        \label{fig:doggo}
    \end{subfigure}
    \caption{Random samples from our \ours{}-Large model.\\ $\alpha=0.2$, $cfg=4.0$ and $24$ steps.}
    \label{fig:viz5}
\end{figure*}

In~\autoref{fig:viz1}, \autoref{fig:viz2}, \autoref{fig:viz3}, ~\autoref{fig:viz4}, and \autoref{fig:viz5}, we present qualitative results from \ours{}-Large. Without any cherry-picking, but with classifier-free guidance (CFG), we demonstrate that our model is capable of generating realistic and diverse images. Maintaining intricate details on both the objects and the background, using only 24 steps.

\end{document}